\documentclass[runningheads]{llncs}
\usepackage{comment}
\usepackage{amssymb} 

\usepackage{amsmath,amsfonts,bm}

\def\eqref#1{equation~\ref{#1}}
\def\Eqref#1{Equation~\ref{#1}}

\def\1{\bm{1}}

\newcommand{\test}{\mathcal{D_{\mathrm{test}}}}

\DeclareMathAlphabet{\mathsfit}{\encodingdefault}{\sfdefault}{m}{sl}
\SetMathAlphabet{\mathsfit}{bold}{\encodingdefault}{\sfdefault}{bx}{n}

\usepackage[algoruled,lined]{algorithm2e} 
\usepackage{algpseudocode}
\usepackage{hyperref}
\usepackage{url}
\usepackage{stmaryrd}
\usepackage{mathtools}
\usepackage{amsfonts}
\usepackage{breqn}
\usepackage{dsfont}
\usepackage{lipsum} 
\usepackage{calc}
\usepackage{graphicx,subfigure}
\usepackage{verbatim}  
\usepackage{floatrow}  

\usepackage{chngcntr}
\usepackage{color}  
\definecolor{gray}{RGB}{180,180,180}
\definecolor{lightgray}{RGB}{230,230,230}
\definecolor{darkgreen}{RGB}{0,40,0}
\newcommand{\remove}[1]{{\color{red}{#1}}}

\newcommand{\notes}[1]{{\textbf{\color{blue}{{#1}}}}}
\newcommand{\draft}[1]{{\color{gray}{{#1}}}}

\usepackage{booktabs} 
\usepackage{multirow}

\begin{document}
\pagestyle{headings}
\mainmatter
\def\ECCVSubNumber{3357}  

\title{Nested Learning for Multi-Granular Tasks} 

\titlerunning{Nested Learning For Multi-Granlar Tasks}
%
\author{Raphaël Achddou\inst{1} \and
J.Matias Di Martino\inst{2} \and
Guillermo Sapiro \inst{2}}
\authorrunning{R. Achddou et al.}
%
\institute{LTCI, Télécom Paris, Institut Polytechnique de Paris, Paris, France \and
Department of Electrical and Computer Engineering, Duke University}
\maketitle

\begin{abstract}
Standard deep neural networks (DNNs) are commonly trained in an end-to-end fashion for specific tasks such as object recognition, face identification, or character recognition, among many examples. This specificity often leads to overconfident models that generalize poorly to samples that are not from the original training distribution. Moreover, such standard DNNs do not allow to leverage information from heterogeneously annotated training data, where for example, labels may be provided with different levels of granularity. Furthermore, DNNs do not produce results with simultaneous different levels of confidence for different levels of detail, they are most commonly an all or nothing approach. To address these challenges, we introduce the concept of \textit{nested learning}: how to obtain a hierarchical representation of the input such that a coarse label can be extracted first, and sequentially refine this representation, if the sample permits, to obtain successively refined predictions, all of them with the corresponding confidence. We explicitly enforce this behavior by creating a sequence of nested information bottlenecks. Looking at the problem of nested learning from an information theory perspective, we design a network topology with two important properties. First, a sequence of low dimensional (nested) feature embeddings are enforced. Then we show how the explicit combination of nested outputs can improve both the robustness and the accuracy of finer predictions. Experimental results on Cifar-10, Cifar-100, MNIST, Fashion-MNIST, Dbpedia, and Plantvillage demonstrate that nested learning outperforms the same network trained in the standard end-to-end fashion.
\end{abstract}

\section{Introduction}
Deep learning is providing remarkable computational tools for the automatic analysis and understanding of complex high-dimensional problems \cite{skin_cancer},\cite{Parkhi_2015}, \cite{breast}. Despite its tremendous value and versatility, methods based on Deep Neural Networks (DNNs) tend to be overconfident about their predictions, and limited to the task and data they have been trained on \cite{Guo_ICML_2017}, \cite{Hein_corr_2018}, \cite{Nguyen_CORR_2014}. This happens among other reasons, because the standard approach to train DNN models consists in optimizing its performance over a specific dataset and for a specific task in an end-to-end fashion \cite{Szegedy_ICLR_2014}. Standard DNNs are not designed to be trained with data of different quality and to simultaneously provide results at multiple granularities. The framework proposed in this work opens the door to this, and in particular when those granular predictions are nested, meaning every subsequent level adds information and has all the information of the previous one (we formalize this concept in the coming section).

\begin{figure}[htp] \label{fig:example1}
    \centering\vspace{-5mm}
    \includegraphics[width=.9\textwidth]{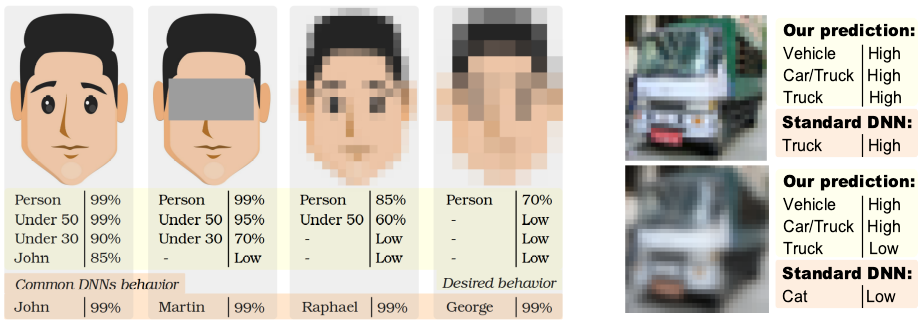}
    \caption{\footnotesize 
    End to end versus nested learning, an illustrative example. On the left, an illustration of a set of nested predictions and their associated confidence given an input image of a face. The top block illustrates a desired behavior. Depending on the quality of the input data, one may be able to provide up to a certain level of nested predictions. The bottom block illustrates how standard DNN-based models often behave when they are trained in a end-to-end fashion. The traditional network is over-confident in its predictions (potentially wrong for the last 3 cases), and provides an all or nothing response instead of responding \textit{only} what it can for the given input quality. On the right, we see a real example with results from the proposed framework; while a sharp image gets all the nested levels with high confidence, a low-quality one gets only confident predictions associated to coarser categories. (Several additional examples on real data are presented in Figure~\ref{app:fig:more_examples_usVSstandard} in the supplementary material.)\vspace{-3mm}
    }
\end{figure}
Consider as an example the case illustrated in Figure~\ref{fig:example1} (left); for high quality facial images, we may be able to infer the person's age group and identity; whereas for poor resolution or occluded examples, only a sub-set of these nested predictions may be achievable. We expect the network to automatically understand what can and cannot be predicted, and this is obtained with the framework proposed in this paper (Figure~\ref{fig:example1}, right). Moreover, nested learning is conceived to leverage training information from diverse datasets, with varying granularity and quality of labels, combining this information into a single model. This will be formalized later in the paper with tools from information theory.

Recently, \cite{Bilal_Corr_2017} showed that convolutional neural networks (CNNs) naturally tend to learn hierarchical high-level features that discriminate groups of classes in the early layers, while the deeper layers develop more specialized feature detectors. 
We explicitly enforce this behavior by creating a sequence of nested information bottlenecks. 
Looking at the problem of nested learning from an information theory perspective, we design a network topology with two important properties.
First, a sequence of low dimensional (nested) feature embeddings are enforced for each level in the taxonomy of the labels.  
This encourages generalization by forcing information bottlenecks \cite{Tishby_corr_2017}, \cite{Tishby_1999}.
Second, skipped connections allow finer embeddings to access information (if available) of the input that may be useful for finer classification but not informative on coarser categories \cite{Unet}.  
Additionally, we show how the explicit calibration and combination of nested outputs can improve the finer predictions and improve robustness.
%
%
The source code associated to this work is open source.\footnote{https://github.com/nestedlearning2019}


The main contributions of this paper are: (1) We introduce the concept of nested learning, where a given level in the hierarchy strictly contains the previous one and strictly adds information; (2) We provide and discuss a structure of deep learning architectures that can be trained (and predict) on data from all levels of a nested taxonomy; we theoretically and empirically study some important properties of this structure in the context of nested learning; (3) We provide a model with multiple outputs, one per level of the nested taxonomy, each one with its own confidence. We show how this design naturally improves the model generalization capabilities and robustness to adversarial attacks.  

\section{Related Work}
The problem of adapting DNNs models and training protocols to encourage nested learning shares similarities with other popular problems in machine learning such as Multi-Task Learning (MTL). 
Though our work is related to MTL because of the similar training challenges, most of the MTL methods tackle the task in a parallel way, without imposing a task nestedness in their architecture \cite{NIPS2016_6393}, \cite{Kokkinos_2017_CVPR},  \cite{ranjan2016hyperface}. 
The idea of learning hierarchical representations to improve classification performance has been exploited prior the proliferation of DNNs, e.g., \cite{Fergus_ECCV_2010},  \cite{Liu_cvpr_2011}, \cite{Zhao_nips_2011}, \cite{zweig_ICCV_2007}. 
Some of these ideas have been incorporated into deep learning methods \cite{Bilal_Corr_2017}, \cite{Deng_ECCV_2014}, \cite{Kim_2018_CVPR}, \cite{Srivastava_nips_2013}, \cite{pmlr-v80-wehrmann18a} and exploited in specific applications \cite{clark2017},  \cite{HL_fashion_classification}, \cite{HL_traffic_recognition}.
The present work and the aforementioned have important differences that will be described at the end of this section.  

Recently, Kim et al.~\cite{Kim_2018_CVPR} proposed a nested sparse network architecture with the emphasis on having a resource-aware versatile architecture to meet (simultaneously) diverse resource requirements. 
A different approach was proposed by Wehrmann et al.~\cite{pmlr-v80-wehrmann18a}, who designed a neural network architecture capable of simultaneously optimizing local and global loss functions to exploit local and global information while penalizing hierarchical violations. 
Triguero et al.~\cite{triguero_2016_JPR} investigated different alternatives to label hierarchical multi-label problems by selecting one or multiple thresholds to map output scores to hierarchical predictions, focusing on performance measures such as the H-loss, HMC-loss and the micro-averaged F-measure. 
Yan et al.~\cite{HDCNN} on the other hand, introduced hierarchical deep CNNs (HD-CNNs) which consists of embedding CNNs into a two-level category hierarchy. They propose to distinguish a coarse class using an initial classifier and then refine the classification into a second level for each individual coarse category. 

Although the works listed above are important, relevant, and related to the ideas here presented, there are notable differences between them and what we propose. 
For example, while \cite{Kim_2018_CVPR} propose a nested architecture providing different (potentially nested) outputs, they do not study how to combine these outputs into a refined single prediction, nor provide a reliable confidence measure associated to them. Furthermore, the architecture they propose has key differences with ours, while they propose a nested hierarchy in an end-to-end fashion (i.e., features associated to the coarse and fine levels are shared from the top to the bottom of the network), we enforce sequential information bottlenecks. 
As we show in the following sections, this sequence of coarse to fine low dimensional representations facilitate a robust calibration and combination of nested outputs, in addition to being a natural way to simultaneously learn with multiple data and label qualities.
\cite{HDCNN} study the problem of nested learning for two nested levels of granularity, and optimize for a final fine prediction. Their design is specific for a two-level category hierarchy while our work generalizes to any number of nested levels, and we can simultaneously train and test up to an arbitrary level of granularity. 
Another important difference is that the focus of their work is on the implementation details and performance of their two-hierarchy network versus traditional end-to-end learning. As they mention in the conclusion of their work, future work should aim to extend their ideas to more than two hierarchical levels and to contextualize their empirical results into a theoretical framework. 
Our work takes steps in these two specific directions. 
This is also a fundamental difference with the works developed by \cite{triguero_2016_JPR} and \cite{pmlr-v80-wehrmann18a}. 
These approaches make the assumption that training samples are annotated for all the granularity levels. In contrast with them, we design a model capable of training (and predicting) on dataset that provide only coarse labels, intermediate levels, or fine labels. 
In contrast with previous work, we show that if testing conditions shift from the ones on training, nested learning can still provide relatively confident coarser labels while avoiding overconfident (erroneous) fine predictions (see, e.g., Figure~\ref{fig:example1}, right; and Figure~\ref{app:fig:more_examples_usVSstandard} in the supplementary material).  
Finally, because our solution can leverage information from datasets with different granularity, we are able to analyze how different proportions of coarse and fine training data affects models cost, performance, and robustness. 

\section{Preliminaries and Notation}
%
Let us use an example to facilitate the definition of the main ideas: assume that we attempt to classify the popular hand written digits of MNIST \cite{lecun-mnisthandwrittendigit-2010}.
An input image can be represented as a realization $x$ of the random variable $X$.\footnote{Upper case letters are used to denote random variables and lower case letters to denote the value of a particular realization.}
We denote as $\mathcal{X}$ the alphabet of $X$.
Associated to each input image $x$, there is a \emph{ground truth label} $y$ that corresponds to the actual number the person writing the character wanted to represent. 
$y$ is a realization of the random variable $Y$, and in this illustrative example, $Y$ can take $10$ different values: $\mathcal{Y}=\{0,1, ..., 9\}$. 
Of course $Y$ and $X$ are not independent random variables and that is why one can infer information about one variable given the other one. 
The problem of classification can be stated as the problem of inferring $y$ from an observed sample $x$, i.e., $Y\rightarrow X \rightarrow \hat{Y}$. 
$\hat{Y}$ denotes a new random variable (estimated from $X$) which \emph{approximates} $Y$. 
More precisely, a common practice is to find a mapping $X \rightarrow \hat{Y}$ such that the probability $P(\hat{Y}\neq Y | X)$ is minimized. 

\vspace{2mm}\noindent\textbf{Nested classification.}
In this work we focus on the inherent nested taxonomy most classification problems have. 
For example, imagine now that we have hand written characters including numbers and letters. 
It would be intuitive to first attempt to classify these characters into two coarse categories: numbers, and letters; and then, depending on this coarse classification we can perform a finer classification, i.e., classifying the numbers into $0-9$ classes, the letters into $a-z$, and so forth. 

\begin{figure}[htp]
\centering
\includegraphics[width=.23\textwidth]{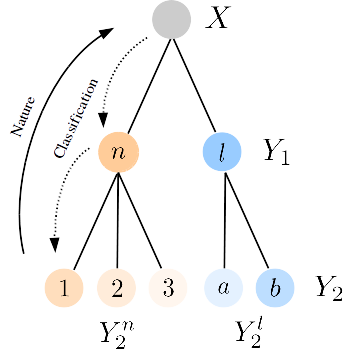}
\hspace{1cm}
\includegraphics[width=.5\textwidth]{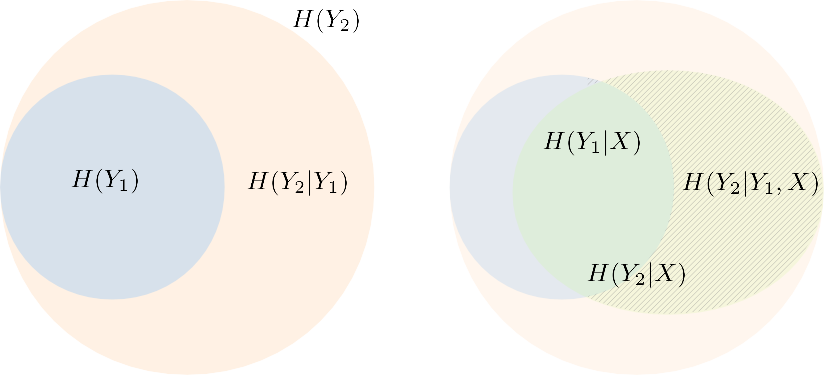}
\caption{\footnotesize
On the left we illustrate the taxonomy of an example of strictly nested labels. First handwritten characters are classified as ``number'' or ``letter,'' and then these categories are refined into specific numbers and letters. $X$, $Y_1$ and $Y_2$ denote random variables representing the input, the coarse label, and the fine label respectively. $Y_i^k$ represents the random variable associated to each value $k$ in the coarser node, i.e., $Y_i$ given that $y_{i-1}=k$, $k\in\mathcal{Y}_{i-1}$.
The diagram in the center, illustrate the entropy of a fine and coarse level, and how having information about a coarser level may reduce the entropy of the fine level. 
The right diagram illustrates the reduction of uncertainty on the labels given the input, and how the uncertainty on the fine labels can be reduced even further if input information and coarse information are combined.}\label{fig:entropy}\vspace{-2mm}
\end{figure}

Of course, we could have an arbitrary number of nested random variables associated to different levels of labels granularity. 
Here, subscripts indicate the granularity of the label, for example, $Y_{i-1}$ is the closest coarse level of $Y_{i}$.
$Y_i^k$ represents the random variable associated to each $k$ value in the closest coarse node, i.e., $Y_i^k$ represents $Y_i$ given that $y_{i-1}=k$, $k\in\mathcal{Y}_{i-1}$.
We now formally define the nested structure, see Figure~\ref{fig:entropy}.
\begin{definition}\label{def:nested_variables}
We define $Y_1, ..., Y_n$ as a discrete sequence of nested labels if $H(Y_i|Y_{i+1})=0$ $\forall i \in [1, n-1]$. $H$ denotes the standard definition of entropy for discrete random variables. 
\end{definition}
\begin{definition}\label{def:stricted_nested}
A discrete sequence of nested labels $Y_1, ..., Y_n$ is strictly nested if $H(Y_i|Y_{i-1})<H(Y_i)$ $\forall i \in [2,n]$. 
\end{definition}

\vspace{-5mm}
\section{Nested Learning}\label{sec:Theory}
The core of this work is to formulate learning problems in a way that the information of the input is extracted in a nested way and allowing a nested structure of predictions. These predictions should work together to improve robustness (as in \cite{pmlr-v80-wehrmann18a}), but simultaneously, be meaningful individually.  As we will show in following sections, this allows us to still provide relatively reliable coarse labels even when the predictions of the finer categories is severely degraded. 

To make this possible we design four main components of the model that work in coordination: a sequence of low dimensional (nested) representations, the output of calibrated predictions, the combination of nested predictions, and finally, a practical and stable training protocol (capable of handling heterogeneously annotated data). The nested sequence of information bottlenecks is inspired by ideas from \cite{Tishby_corr_2017} and is described in detail in Section~\ref{subsec:HierarchicalInformation}. Forcing this sequence of low dimensional representation has also the advantage of making numerically tractable the calibration of individual outputs, as later described in Section \ref{subsec:Scores_calibration}. Then, we show that individual outputs can be used independently or in combination, as described in Section~\ref{subsec:CombinationOfOutputs}. Since training models with heterogeneous data is challenging \cite{Kokkinos_2017_CVPR} we discuss in Section~\ref{subsec:training_nested} how to solve some of these challenges. Finally, we present experimental analysis in Section \ref{sec:Experiments}. 
\subsection{Nested information bottlenecks and the role of skipped connections} \label{subsec:HierarchicalInformation}
\begin{figure}[htp]
\centering\includegraphics[width=.95\textwidth]{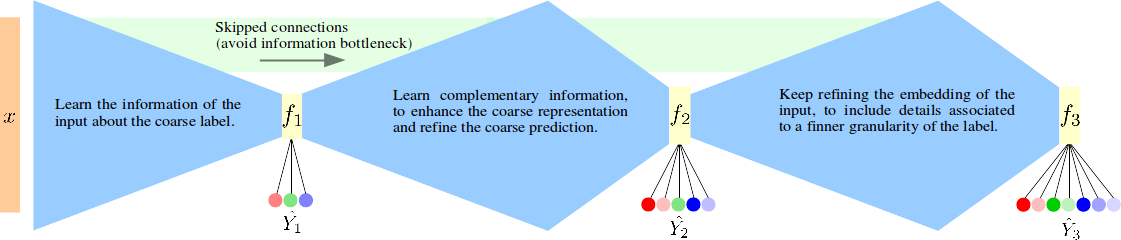}
\caption{\footnotesize Illustrative scheme of the proposed framework. 
From left to right, the input data $x\sim X$, a first set of layers that extract from $X$ a feature representation $f_1$, which leads to $\hat{Y}_1$ (estimation of the coarse label $Y_1$). 
$f_1$ is then jointly exploited in addition with complementary information of the input. 
This leads to a second representation $f_2$ from which a finer classification is obtained. 
The same idea is repeated until the fine level of classification is achieved. 
It is important to highlight that this high level description of the proposed model can be implemented in multiple ways as we further discuss in the following sections.}\label{fig:framework}\vspace{-2mm}
\end{figure}

Assume the input $X$ has information about a sequence of strictly nested labels $Y_i$, i.e., $H(Y_i|X) < H(Y_i)$, $H(Y_i|X) > H(Y_{i-1}|X)$, and $H(Y_i|X) < H(Y_{i}|Y_{i-1},X)$, as illustrated in Figure~\ref{fig:entropy} and defined in the previous section.
Exploiting these inequalities, we will sequentially \emph{compress} the information on $X$ using standard DNN layers (convolutional, pooling, normalization, and activation) as we schematically illustrate in Figure~\ref{fig:framework}.
We begin by guiding the network to find a low dimensional feature representation $f_1$ such that $H(f_1(X))\ll H(X)$ while, $I(f_1(X),Y_1)$ is \emph{close} to $I(X,Y_1)$. $I(U,V)$ stands for the standard mutual information between discrete random variable $U$ and $V$. 
(DNNs are remarkably efficient at compressing and extracting the mutual information between high dimensional inputs and target labels \cite{Tishby_corr2_2017}, \cite{Tishby_corr_2017}.) 

The second step consists in learning the \emph{complementary} information such that when combined with the representation $f_1$, it allows to achieve a second representation $f_2$ from which the second hierarchical label $Y_2$ can be inferred. 
To this end, skipped connections play a critical role as we will discuss next. 

Using the definition of mutual information and the property that the sequence $Y_i$ is a set of strictly nested labels, we have that
\begin{equation}\label{eq:information_bottleneck}
    I(X,Y_i) = H(X) - \underbrace{H(X|Y_i)}_{> H(X|Y_{i+1})} < I(X,Y_{i+1}).
\end{equation}
On the other hand, we want each feature embedding $f_i$ to compress the information of $X$ while $I(f_i(X), Y_i) \approx I(X, Y_i)$.
\Eqref{eq:information_bottleneck} implies that the finer the classification the more information from $X$ is required.

Notice that while in most DNNs architectures skipped connections are included to encourage the model compactness and to mitigate vanishing gradients \cite{szegedy2017inception}, in the present work they are included for a more fundamental reason. 
If we do not consider skipped connections, $X\rightarrow f_i(X) \rightarrow f_{i+1}(X)$ forms a Markov chain where $I(X, f_{i+1}(X)) \leq I(X,f_{i}(X))$ (data-processing inequality), this would contradict \Eqref{eq:information_bottleneck}.

We will validate these ideas in Section \ref{sec:Experiments} . First we study the accuracy and robustness of for different networks architectures. Then, we measure approximations of the mutual information \cite{belghazi2018mine} between different feature representations along the network, with and without skipped connections, observing that in fact the information flow is twice as large when skipped connections are included.

\subsection{Combination of nested outputs} \label{subsec:CombinationOfOutputs}
We will present in the following sections experimental evidence showing that nested learning leads to an improvement in performance and robustness. 
The explicit combination of nested predictions can improve the accuracy and robustness even further.
To this end, we refine the fine prediction leveraging the information of all the coarser outputs, i.e., $\{\hat{Y}_1, ..., \hat{Y}_i\}\rightarrow\tilde{Y}_i$.

Let us define $s_i(q)$ the network output score associated to the event $Y_i=q$. 
In general, if $s_i(q)>s_i(w)$ most likely $P(Y_i=q)>P(Y_i=w)$, but $P(Y_i=q) \neq s_i(q)$. 
In other words, a score value of $0.3$ does not mean the sample belongs to this class with probability equal to $30\%$.
This mismatch can be addressed by calibrating the score outputs, which consists of mapping output scores into an estimation of the class probability $s_i(q) \rightarrow P_{\hat{Y_i}}(q)$. The problem of calibration is well defined and thoroughly explained in \cite{zadrozny2002transforming}. 
Let us denote $P_{\hat{Y_i}}(q)$ the calibrated output of the network that approximates $P(Y_i=q)$.
(We address how calibration is performed in the following section.)
Then, we can use the estimated probability associated to a fine label $P_{\hat{Y}_i}$ to compute the conditional probability $P(Y_{i} = y_i | Y_{i-1} = k)$.
This is achieved by re-normalizing the finer labels associated to the same coarse label, i.e.,
\begin{equation}\label{eq:normalization_outout}
P_{\hat{Y}_i|\hat{Y}_{i-1}}(q) = \frac{P_{\hat{Y}_i}(q)}{\sum_{w\in \mathcal{Y}_i^{k_q}}P_{\hat{Y}_i}(w)},
\end{equation}
where $\mathcal{Y}_i^{k_q}$ denotes the set of labels at granularity level $i$ that share with $q$ the same coarser label $k_q$. 
Finally, the estimated conditional probability is combined with the prior of the coarser prediction to recompute the fine prediction $P'_{\hat{Y}_i}(q) = P_{\hat{Y}_i|\hat{Y}_{i-1}}(q)P_{\hat{Y}_{i-1}}(k_q)$, which is then refined recursively until we reach the coarser level: $P'_{\hat{Y}_i}(q) = P_{\hat{Y}_i|\hat{Y}_{i-1}} P_{\hat{Y}_{i-1}|\hat{Y}_{i-2}}\, ...\, P_{\hat{Y}_0}$. This is a generalization of the combination method proposed for two nested levels by Yan et al. \cite{HDCNN}.

Related alternatives are presented by Kuncheva \cite{kuncheva_2004} in the context of combining outputs of ensembles of classifiers. However, those methods are agnostic to the labels' taxonomy.
Experimentally, we observed that the different combination methods perform similarly on test data that matches the training distribution, while the proposed method outperforms the others when test samples are distorted.
We later compare some of those combination methods (e.g., Mean, Product, and Majority Vote) to the one that we designed specifically for multiple nested outputs. (Details and numerical results are provided in the supplementary material, Section \ref{app:sec:combination} and Table~\ref{app:tab:combinationsMNIST101010}.)
\vspace{-2mm}
\subsection{Training}\label{subsec:training_nested} 
\vspace{-1mm}
Let $G_{\theta,\eta}(x) = (f_{i}(x,(\theta_j)_{j=1,..,i}), g_i(f_i,\eta_i))_{i=1,..,m}$ be the function coded by our network, where $m$ denotes the number of granularity levels and as before $x$ represents an input sample.
Each sub-function $g_i$ corresponds to the output of granularity $i$ (computed from the feature bottleneck $f_i$). 
$G$ depends on parameters $(\theta_j)_{j=1,..,i}$ which are common to the sub-functions of coarser granularities, and some granularity-specific parameters $\eta_i$. 
The general framework of the architecture follows Figure~\ref{fig:framework}, meaning a trunk of convolutional filters with parameters $\theta$ and fully connected layers for each intermediate outputs with parameters $\eta$.

Training this type of model with a disparity of samples per granularity is challenging, and naively sampling random batches of training data leads to a noisy gradient computation \cite{Kokkinos_2017_CVPR}. 
In order to overcome this issue, we organize the training samples and train the network in a cascaded manner.
The dataset $\mathcal{D}$ is organized in subsets of samples labeled up to granularity $i$ for $i=1,..,m$. 
Formally we can write $\mathcal{D} = (\bf{x}, \bf{y})$ with $\bf{x}$ the set of inputs and $\bf{y}$ the set of labels.
We consider that $\bf{x} = (\bf{x}_i)_{i=1,..,m}$ and $\bf{y} = (\bf{y}_i)_{i=1,..,m}$, where $\mathcal{D}_i = (\bf{x}_i, \bf{y}_i)$ represents the subset of data for which the label is known up to the granularity level $i$.

Having the dataset partitioned in this fashion naturally leads to a cascaded training of the network. 
We train the model to solve a sequence of optimization problems using $(\bf{x}_i,\bf{y}_i)$ as the training examples at each step. 
We can write this sequence as:
$ (P_i) : \min_{(\theta_j,\eta_j)_{j=1,..,i}} \sum_{j=1}^{i}\alpha_j \mathcal{L}_{n_j}(\hat{Y}_{j},Y_j), $
where $\mathcal{L}_{n}$ is the $n$-categorical cross-entropy and $\alpha$ are the weights for each individual loss.
We start by training the network on coarse labels, and we gradually add finer labels optimizing deeper parameters in an iterative way. Experimental results in Section \ref{sec:Experiments} support this (also) nested training approach.


\subsection{Scores calibration}\label{subsec:Scores_calibration}
As shown by \cite{Hein_corr_2018}, deep neural networks with ReLU activations tend to produce overconfident results for samples which are out of the training distribution. 
In order to mitigate this issue and obtain a more accurate estimation of the classes probabilities we consider a two step calibration method.
The first step consists in adding a ``rejection'' class for each level of granularity. 
Synthetic samples associated to this class are generated from an uniform distribution.
By training the network with this supplementary class we mitigate the problem of overconfidence for samples that are far from the training distribution.

Keeping a fixed coverage on the input space would require a number of synthetic samples that grows exponentially with the number of dimensions. 
We solve this practical problem by injecting the samples associated to the rejection class at the low-dimensional bottleneck representation associated to each classification level. 
This means that, inherently, our regularization works in a multi-scale fashion in the taxonomy of predictions. 
We illustrate this idea on a simple 1D toy example in the supplementary material (Section~\ref{app:sec:implem_details}). 

The second calibration step is a temperature scaling introduced in \cite{Guo_ICML_2017}.
This technique consists in scaling the output of the fully-connected layer before the softmax activation by an optimal temperature parameter.
Given $x$ the input data and $g$ the function coded by a neural network before the softmax activation $\sigma$, the new calibrated output is given by $\bar{g} = \sigma(\frac{g}{T})$.
The temperature parameter $T$ is tuned so that the mean confidence of the predictions matches the empirical accuracy; more precisely, we want to minimize $\mathbb{E}\left[|P(\hat{Y}=Y|\hat{p}=p)-p|\right]$, where as before, $\hat{Y}$ denotes the network prediction of $Y$, and $\hat{p}$ is the empirical confidence associated to it. 
The previous expression can be approximated, using a partition of the validation data, by computing the Expected Calibration Error (ECE) $ECE = \sum_{j=1}^{N}\frac{|B_j|}{n}|acc(B_j)-conf(B_j)|$ \cite{naeini2015_AAAI}.
This measure takes the weighted average between the accuracy and confidence on $N$ bins $B_j$, $j=1...N$. 
Here, $n$ denotes the total number of samples, and $|B_j|$ the number of samples on the bin $B_j$. 
\vspace{-2mm}
\section{Experiments and Discussion}\label{sec:Experiments}
We consider six publicly available datasets for experimental evaluation: the handwritten digits from MNIST \cite{lecun-mnisthandwrittendigit-2010}, the small clothes images from Fashion-MNIST \cite{xiao2017/online}, CIFAR10 \cite{cifar10}, CIFAR100 \cite{cifar100}, the Plantvillage dataset \cite{plantvillage2016}, and DBPEDIA \cite{auer2007dbpedia} made of articles of Wikipedia.\footnote{ https://www.kaggle.com/danofer/dbpedia-classes}
We created visually based taxonomies for the first three datasets, and used the provided taxonomy for the remaining ones. Details on the taxonomy of each dataset are given in the supplementary, Section~\ref{app:sec:label_grouping}.
We first compare end-to-end training to our proposed nested learning approach.
We assess accuracy on \emph{clean} data as well as robustness to different types of distortions and adversarial attacks. 
We also evaluate the impact of skipped connections, the training protocol, the training budget, and the network architecture.
Additional experiments are presented in the supplementary material, sections~\ref{app:sec:more_examples} and \ref{app:sec:additional_experiments}, and implementation details (architecture, training, etc) are provided in Section~\ref{app:sec:implem_details}.

\vspace{2mm}\noindent\textbf{Does coarsely annotated data improve the prediction of the fine task?} %
Let us define $|\mathcal{D}_i|$ as the number of training samples that are annotated up to the level of granularity $i$. (Since we are working with strictly nested classes, knowing the label for the level $i$ provides all the coarser labels $j$ for $j<i$.)
To understand how coarse annotations impact the performance on a finer task, we compared models trained exclusively with fine data $\mathcal{D}_A \equiv \mathcal{D}_3$ and models trained with the same amount of fine data plus coarse data $\mathcal{D}_B \equiv \mathcal{D}_3 \cup \mathcal{D}_2 \cup \mathcal{D}_1$.

The results are presented in Table~\ref{tab:cifar10_202020vs20} for CIFAR10; in Table~\ref{tab:real_world_data} for Dbpedia, Plantvillage, and Cifar100 datasets; and in Table~\ref{app:tab:mnist_fmnist_202020vs20} in the supplementary for MNIST and F-MNIST. 
Compare for instance the results shown in Table~\ref{tab:cifar10_202020vs20}, and in particular, row G with respect to row I. The former shows the results of the model trained exclusively with fine annotations ($|\mathcal{D}_3|= 10^4$), while the latter corresponds to the model trained with additional coarse and middle annotations ($|\mathcal{D}_{i}|=10^4$ for $i = 1,2,3$). Naturally, training with additional coarse and middle data improves the accuracy of the coarse and middle prediction. More interestingly, it also improves the robustness of the fine label (see how the accuracy increases and the overconfidence decreases), in particular, when testing on samples that slightly differ from the ones at training.\footnote{Distortions 1 to 4 in Table~\ref{tab:cifar10_202020vs20} correspond to four levels (increasing the severity) of ``turbulence-like'' image distortion, the implementation of this distortion is inspired by~\cite{ipol_turbulence} (details are provided in the supplementary material, Section \ref{app:sec:perturbation}).} This observation is consistent with the results obtained when testing the models response to adversarial attacks, as we shall discuss in the following experiments. 

\vspace{2mm}\noindent\textbf{With a fixed training budget, what is the better trade-off?}
The previous results suggest that having additional coarse annotations improves the model capacity to predict the coarse labels, as well as the performance and robustness of the finer tasks. We now study for a specific budget (this is, getting coarse annotations comes at the expense of less fine annotations), which are the level of annotations that contribute the most to improve the learning process?

%
Looking at tables~\ref{tab:cifar10_202020vs20} and~\ref{tab:real_world_data} we see that the model trained with additional coarse and middle samples tends to be more robust and less overconfident, even compared to models trained with a larger amount of fine annotations. 
These results were replicated across different domains, for instance, they include: the classification of images of characters, fashion and natural images as well as text documents. 
Complementing these results, Section~\ref{app:sec:MTL} in the supplementary material compares the proposed nested architecture with a multi-task learning approach.
\begin{table}[t]
\begin{center}
\hspace{-5mm}
    \scriptsize
    \begin{tabular}{|c|c||c|c|c|c|c|}
    \hline
        ind & method & Original & Distortion 1 &  Distortion 2 &  Distortion 3 &  Distortion 4 \\
        \hline
        \hline
         A&Coarse (end-to-end), $|\mathcal{D}_3| = N$ &       96.0 / 97.6 &   87.0 / 94.0 &          82.5 / 93.4 &          80.2 / 92.9  &          77.0 / 93.0  \\
         B&Coarse (end-to-end), $|\mathcal{D}_3| = \frac{3N}{2}$ & \textbf{96.8} / 98.2 &   86.2 / 93.9 &  82.1 / 93.1 &  78.3 / 92.7  &  75.1 / 92.9  \\
          C&Coarse (nested), $|\mathcal{D}_{1,2,3}| = N$&       96.5  / 96.7  & \textbf{87.8} / 92.5&          \textbf{84.9}  / 91.5&          \textbf{81.4} / 90.9 &          \textbf{78.2} / 90.6  \\
          \hline
          \hline
        D&Middle (end-to-end), $|\mathcal{D}_3| = N$&        84.1 /  89.8 &  65.2 / 79.6 &          56.7 /  76.0 &          48.9 / 74.9  &          42.6 / 75.9     \\
        E&Middle (end-to-end), $|\mathcal{D}_3| = \frac{3N}{2}$&        \textbf{87.5} /  92.8 &  \textbf{65.5} / 81.5 &          56.3 /  79.0 &          47.8 / 78.7  &          41.3 / 79.4 \\
         F&Middle (nested), $|\mathcal{D}_{1,2,3}| = N$&          85.2 / 85.0 & 65.4 / 73.5 &          \textbf{58.1} / 70.4 &          \textbf{50.3} / 69.5 &          \textbf{43.9}  / 69.7 \\
          \hline
          \hline
         G&Fine (end-to-end), $|\mathcal{D}_3| = N$&          75.9 / 88.4 & 50.3 / 67.9 &          42.8 /64.9 &          34.2 / 65.5  &          28.4 /  73.4     \\
         H&Fine (end-to-end), $|\mathcal{D}_3| = \frac{3N}{2}$&          \textbf{81.0} / 88.4 & \textbf{52.0} / 73.3 &          41.8 /71.4 &          32.8 / 71.9  &          26.8 /  73.4     \\
          I&Fine (nested), $|\mathcal{D}_{1,2,3}| = N$&          77.4 /  77.7 & 51.6 / 62.6&          \textbf{43.2} /59.8  &          \textbf{34.7} / 57.7&          \textbf{29.1} / 57.7  \\
          \hline
\end{tabular}
\end{center}
    \caption{\footnotesize
    Accuracy and mean confidence ($Acc\% / Conf\%$) for Cifar10. Coarse, fine, and middle indicate the accuracy at each level. End-to-end denotes the model trained exclusively for the fine task, nested denotes the same architecture trained to learn coarse, middle, and fine predictions. We set $N = 10^4$.
    We repeated each experiment 10 times and report the mean across these repetitions. We also computed the standard deviation across the results which was in all the cases below $0.5$. }\vspace{-2mm}
    \label{tab:cifar10_202020vs20}
\end{table}
\begin{table}[htp]
    \centering
    \scriptsize
    \begin{tabular}{|c||c|c|c||c|c|c||c|c|}
    \hline
    Method & \multicolumn{3}{c||}{DBPEDIA}  &  \multicolumn{3}{c||}{Plantvillage} & \multicolumn{2}{c|}{CIFAR100}  \\
    \hline
    \hline
    \multirow{2}{*}{Amount of samples}&Coarse&Middle&Fine&Coarse&Middle&Fine&Coarse&Fine\\
     &50000&50000&50000&4500&4500&9000&25000&25000\\
    \hline
        End-to-end &91.8&81.8&76.4&95.2&94.5&92.3 &70.0& 59.6  \\
        Nested &\textbf{98.1}&\textbf{93.4}&\textbf{84.7}& \textbf{97.9}&\textbf{97.5}&\textbf{94.6}& \textbf{79.3} & \textbf{64.9}  \\
    \hline
    \end{tabular}
    \caption{Accuracy of the end-to-end and nested models for three challenging datasets: Dbpedia \cite{auer2007dbpedia}, Plantvillage \cite{plantvillage2016} and Cifar100 \cite{cifar100}. In the first row we report the amount of coarse, middle, and fine data that was used for the nested training. The amount of data used for the end-to-end training corresponds to the amount of fine data.}
    \label{tab:real_world_data}
\end{table}

\vspace{1mm}\noindent\textbf{Robustness to adversarial attacks.} In the previous experiments we tested how nested learning can improve the accuracy and robustness for different models on different tasks. In particular, we tested how sensitive the models are to image morphological distortions.  
To complement the previous experiments, we now compare how difficult is to fool nested and end-to-end models with active (e.g., adversarial) attacks. 
The principle of adversarial attacks is to add a well fitted noise to a sample $x$, such that the model's prediction becomes incorrect. 
Most of the white-box attacks -where the attacker knows everything about the model- are gradient based.
For example, the Fast Gradient Sign Method (FGSM) \cite{goodfellow2014adversarialattack} can be formulated as $x_{adv} = x + \epsilon \text{ sign}(\nabla_{x}(L(f(x), y_{true}))),$ 
where $f$ is the function coded by the network, $y_{true}$ is the true label of sample $x$, $\epsilon$ is the magnitude of the adversarial noise, and $L$ is the loss associated to the prediction we are misleading.
We tested some of the most popular state-of-the-art gradient-based attacks such as Deepfool \cite{moosavi2016deepfool} and Saliency based attacks \cite{papernot2017practical}. 

As we can see in Table \ref{tab:adv1}, to reach a given error rate, the attacker needs to add more than twice as much adversarial noise to the network trained in a nested fashion compared with its standard counterpart. 
\begin{table}[htp]
    \centering
    \scriptsize
    \begin{tabular}{|c||c|c|c|c|}
        \hline
          Method & $\epsilon@$ER=5\% & $\epsilon@$ER=10\% & $\epsilon@$ER=15\% & $\epsilon@$ER=30\% \\
        \hline
        end-to-end & $6.0\times 10^{-3}$& $1.0\times 10^{-2}$& $1.4\times 10^{-2}$& $2.2\times 10^{-2}$\\
        nested     & $1.4\times 10^{-2}$& $2.2\times 10^{-2}$& $3.0\times 10^{-2}$& $5.0\times 10^{-2}$\\
        \hline
    \end{tabular}
    \caption{Amplitude of the perturbation required to achieve a given error rate. ER stands for the error rate, this is, $\epsilon@\text{ER}=X\%$ denotes the magnitude $\epsilon$ required to mislead $X$ percent of the test samples. In this experiments the FGSM adversarial attack technique is evaluated over the MNIST dataset.}
    \label{tab:adv1}
\end{table}
Moreover, as we report in Table~\ref{tab:adv2}, when we fit an attack on the fine output, the coarse and intermediate outputs are less affected for models learned with nested learning. This can be explained intuitively because our model is learning a feature representation with a hierarchical nested structure as we show and discuss in the supplementary material Section~\ref{app:Geometry}. It is important to highlight that our model is not explicitly designed to overcome adversarial attack (contrary to, e.g., \cite{madry2017towards}, \cite{samangouei2018defense}, and \cite{sinha2017certifying}), rather, because the feature representation has an underlying nested structure, the learned models are inherently more robust to adversarial attacks. 
\begin{table}[htp]
    \centering
    \scriptsize
    \begin{tabular}{|c||c|c|c|}
        \hline
        Type of attack & coarse acc. & middle acc. & fine acc. \\
        \hline
        \hline
        FGSM nested          &\textbf{86.8}&\textbf{60.9}&0.0\\
        FGSM end to end      &25.2&15.6&0.0\\
        \hline
        Deepfool nested      &\textbf{57.2}&\textbf{35.7}&0.0\\
        Deepfool end to end  &44.5&27.5&0.0\\
        \hline
        Saliency nested      &\textbf{82.2}&\textbf{71.5}&0.0\\
        Saliency end to end  &27.4&17.2&0.0\\
        \hline
    \end{tabular}
    \caption{Accuracy for the middle and coarse prediction when the fine prediction is adversarially attacked. In contrast with the results reported in Table~\ref{tab:adv1}, in this experiment we increased for each test sample the magnitude of the attack ($\epsilon$) until its fine prediction becomes incorrect. Then, we compute the middle and coarse prediction for the end-to-end and nested models. The nested model explicitly provides middle and coarse outputs, while for the end-to-end model the nested and coarse labels are computed from the fine prediction.}
    \label{tab:adv2}
\end{table}

\begin{figure}[h]
    \includegraphics[width = .45\textwidth]{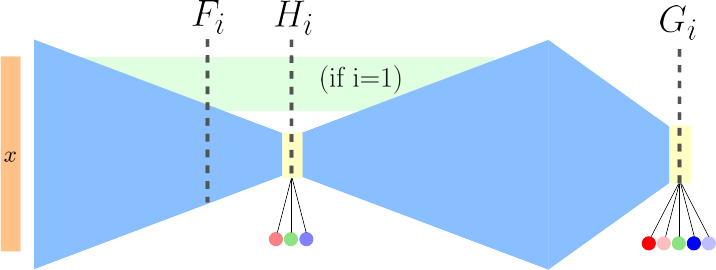}
    \includegraphics[width = .45\textwidth]{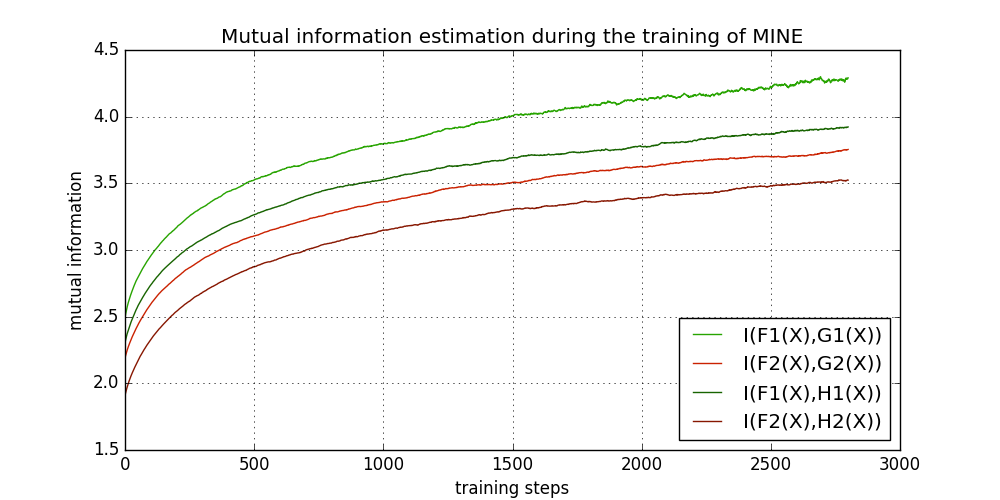}
    \caption{Empirical evaluation of the information flow with and without skipped connections. Left: sketch of the measured sections for (the sub-index $i=1$ represents the quantities of the model with SC, and $i=2$ the model without SC). Right: evolution of the estimated mutual information (MI) for $I(F_1(X),G_1(X))$, $I(F_2(X),G_2(X))$, $I(F_1(X),H_1(X))$, and $I(F_2(X),H_2(X))$ during the optimization of the MINE estimator \cite{belghazi2018mine}. Implementation details are provided in the supplementary material Section~\ref{app:sec:skipedConnections}.}
\label{fig:MI}
\end{figure}

\vspace{-6mm}\noindent\textbf{Information bottlenecks and the role of skipped connections.}
Skipped connections (SC) are included in order to allow information to flow from the input to the finer feature representation (avoiding the \emph{data processing} inequality as we discussed in Section~\ref{subsec:HierarchicalInformation}). To test how the ideas outlined in Section~\ref{subsec:HierarchicalInformation} affect deep models, we compared equivalent models with and without SC, and we empirically measured the flow of information as described in the supplementary material (Section~\ref{app:sec:skipedConnections}). In Table~\ref{tab:MNIST_impact_of_skiped_connections} we compare the accuracy of the model with and without SC (on \emph{clean} and distorted test samples), and in Figure~\ref{fig:MI}, we illustrate the empirical estimation of the flow of information for both models (in the figure, $i=1$/$i=2$ denotes the model with/without SC). 
\begin{table}[htp]
\begin{center}
\vspace{-2mm}
\scriptsize
 \begin{tabular}{|c||c|c|c|c|c|} 
 \hline
  Method & Original & Distortion 1 & Distortion 2 & Distortion 3 & Distortion 4 \\ [0.5ex] 
 \hline
 \hline
 Coarse (without SC) &      99.6 & 96.3 & 91.3 & 85.3 & 79.2\\ 
 Coarse (with SC) Ours &      \textbf{99.7} & \textbf{97.2} & \textbf{93.2} & \textbf{87.7} & \textbf{81.4}\\
 \hline
 Middle (without SC) &      99.3 & 92.1 & 79.4 & 66.8 & 58.4\\ 
 Middle (with SC) Ours &      \textbf{99.5} & \textbf{95.1} & \textbf{87.3} & \textbf{79.2} & \textbf{65.5}\\ 
 \hline
 Fine (without SC)   &      98.9 & 88.2 & 72.4 & 56.4 & 44.0\\
 Fine (with SC) Ours  &      \textbf{99.2} & \textbf{94.2} & \textbf{84.9} & \textbf{69.5} & \textbf{53.2}\\ 
 \hline
\end{tabular}
\end{center}
    \caption{\footnotesize Comparison of the same network structure, trained on the same coarse, middle, and fine data, with and without SC. 6000 of fine, middle and coarse samples of MNIST dataset where selected for training. As in the previous experiments, Distortion 1-4 correspond to test distorted samples with turbulence-like distortion (described in the supplementary material, Section~\ref{app:sec:perturbation}).}\vspace{-4mm}
    \label{tab:MNIST_impact_of_skiped_connections}
\end{table}

We observe that the model with SC performs slightly better on images from the original distribution, and much better on distorted data. 
We also see that the performance gap on the coarse prediction is relatively small, while as expected, the gap increases for the fine and middle prediction.
In addition, as we illustrate in Figure~\ref{fig:MI} and quantitatively describe in the supplementary material (Section~\ref{app:sec:skipedConnections}), the difference in the mutual information between coarse and fine feature embeddings is duplicated when SC are removed. In other words, when SC are included, the relative flow of information from the input to the deeper embeddings increases by a factor of two. 
\vspace{2mm}\noindent\textbf{Training with nested data.}
As we discussed in Section~\ref{subsec:training_nested}, training DNN models with heterogeneous data is challenging. We experimentally compared the proposed cascaded methodology with the standard methodology (i.e., selecting batches of random samples out of the training set).
As samples are annotated heterogeneously (with all or some fine to coarse annotations), we defined the unified loss in each mini-batch as 
$
\mathcal{L}_{batch} = \sum_{j=1}^M\sum_{i=1}^N \alpha_i\times\omega_{ij}\mathcal{L}_i(y_{ij},\hat{y_{ij}}),
$
where $M$ denotes the number of samples in the batch, $N$ the number of nested levels, $\omega_{ij}$ is a masking vector indicating if the sample $j$ is annotated or not for the level of granularity $i$, and $\alpha_i$ are global weights inversely proportional to the amount of annotated data for each level $\mathcal{L}_i$. 

\begin{figure}[ht]
    \centering
    \includegraphics[width = 0.32\textwidth]{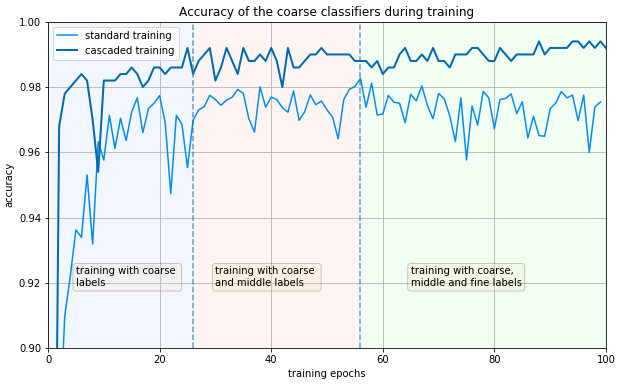}
    \includegraphics[width = 0.32\textwidth]{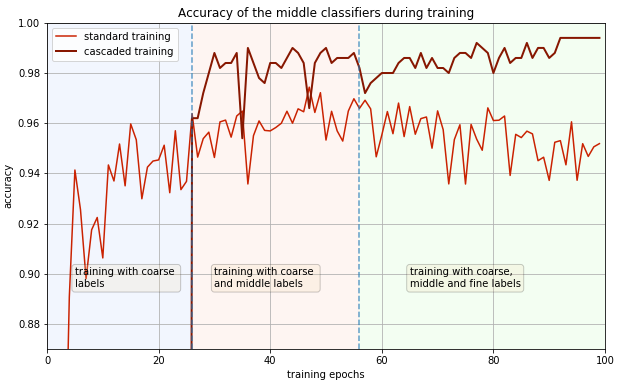}
    \includegraphics[width = 0.32\textwidth]{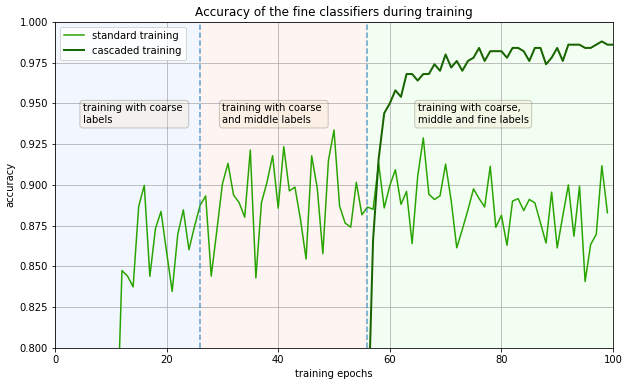}
    \caption{End-to-end versus nested training. We compare the accuracy of the same model trained with a cascaded and a traditional training scheme. Left, evolution of the accuracy of the coarse prediction; center, accuracy of the intermediate prediction; and right, the fine prediction.}
    \label{fig:casc_vs_trad}
\end{figure}

Empirical results are provided in Table \ref{tab:cascadedvstraditional}, we observe that cascaded training achieves substantially better performances than the traditional approach. 
The gaps are very significant also on the training distribution as we observe in the first column.
Additionally, we studied the behaviour of the network during training, e.g., see the results presented in Figure \ref{fig:casc_vs_trad}. 
As we can see, the proposed protocol is more suitable for the proposed multi-level problem leading to faster convergence and better models.  
After 26 epochs of coarse learning, we add the intermediate data for 30 epochs before adding the fine data for 44 more epochs. 
In addition we observe that the accuracy of the cascaded training classifiers oscillates significantly less than the accuracy of the traditionally trained classifiers. 
This provides further evidence that cascade training mitigates the noises of the stochastic gradient estimation.
\vspace{-5mm}
\begin{table}[htp]
\scriptsize
\centering
        \begin{tabular}{|c||c|c|c|c|c|c|}
        \hline
                  Method & Original & Distortion 1 &  Distortion 2 &  Distortion 3 &  Distortion 4 \\
        \hline
             Coarse (traditional),$|\mathcal{D}_{1,2,3}| = N$  & 98.8 / 82.0 &         94.9 / 75.4 &          88.4 / 72.5 &          80.2 / 71.0 &          73.4 / 70.8      \\
             Coarse (cascaded), $|\mathcal{D}_{1,2,3}| = N$ &   \textbf{99.5} / 99.5 &       \textbf{96.9} /  97.5&          \textbf{92.3} / 95.4&          \textbf{85.9} / 93.8&          \textbf{79.6} / 92.1      \\
             \hline
             Middle (traditional), $|\mathcal{D}_{1,2,3}| = N$ & 96.9 /  81.2 &         91.1 / 74.2&          81.1 / 69.5&          66.8 / 67.6 &          53.5 / 69.3    \\
             Middle (cascaded), $|\mathcal{D}_{1,2,3}| = N$  & \textbf{99.2} / 99.2   &      \textbf{95.5} / 96.0 &          \textbf{87.9} / 92.6 &          \textbf{77.0} / 89.4 &          \textbf{66.5} / 87.9 \\
             \hline
             Fine (traditional), $|\mathcal{D}_{1,2,3}| = N$ &   94.2 / 84.1 &       86.7 / 79.0&          73.4 / 71.9&          53.2 /  68.8 &          37.6/ 70.5      \\
              Fine (cascaded), $|\mathcal{D}_{1,2,3}| = N$ &     \textbf{98.6}  / 98.4 &     \textbf{92.8} /  93.5&          \textbf{82.8} / 87.8 &          \textbf{67.5} / 83.0  &          \textbf{50.6} / 80.4    \\
              \hline
        \end{tabular}
    \caption{\footnotesize Accuracy and mean confidence ($Acc\% / Conf\%$) for the MNIST dataset. Coarse, fine, and middle indicate the accuracy at each level of the label. We compare the results of a traditional training and cascaded training on the same architecture. In this experiment we set $N = 6000$.}
    \label{tab:cascadedvstraditional}
\end{table}
\vspace{-10mm}
\vspace{-10mm}
\section{Conclusion}\label{sec:Conclusion}
We introduced the concept of {\it nested learning}, providing some general guidelines and architecture notions to handle heterogeneous data and design systems with nested predictions. 
We showed that such models are more flexible, perform better, and are more robust to both distorted data and active adversarial attacks. 
Moreover, the proposed framework allows to leverage information from datasets annotated with different levels of granularity. 
Additionally, experiments suggest that nested models only gradually break as the quality of the test data deteriorates. 
We showed that implementing nested learning using a hierarchy of information bottlenecks provides a natural framework to enforce calibrated outputs, where each level comes with its own confidence value. 
%
We further demonstrated that if the amount of fine training samples is constant, then adding samples with only coarse annotations increases the performance and robustness for the fine task. 

To recap, the introduced nested learning framework performs as expected from our own human experience, where for good data we can provide high level inference with high confidence; and when the data is not so good, we can still provide with high confidence some level of inference on it.

%
%

\clearpage
%
%
\bibliographystyle{splncs04}
\bibliography{egbib}

\begin{thebibliography}{10}
\providecommand{\url}[1]{\texttt{#1}}
\providecommand{\urlprefix}{URL }
\providecommand{\doi}[1]{https://doi.org/#1}

\bibitem{auer2007dbpedia}
Auer, S., Bizer, C., Kobilarov, G., Lehmann, J., Cyganiak, R., Ives, Z.:
  Dbpedia: A nucleus for a web of open data. In: The semantic web, pp.
  722--735. Springer (2007)

\bibitem{Behrisch:2016:MRM:3071534.3071595}
Behrisch, M., Bach, B., Henry~Riche, N., Schreck, T., Fekete, J.D.: Matrix
  reordering methods for table and network visualization. Comput. Graph. Forum
  \textbf{35}(3),  693--716 (Jun 2016)

\bibitem{belghazi2018mine}
Belghazi, M.I., Baratin, A., Rajeswar, S., Ozair, S., Bengio, Y., Courville,
  A., Hjelm, R.D.: {MINE}: Mutual information neural estimation. arXiv preprint
  arXiv:1801.04062  (2018)

\bibitem{Bilal_Corr_2017}
Bilal, A., Jourabloo, A., Ye, M., Liu, X., Ren, L.: Do convolutional neural
  networks learn class hierarchy? IEEE Transactions on Visualization and
  Computer Graphics  \textbf{24}(1),  152--165 (2018).
  \doi{10.1109/TVCG.2017.2744683}

\bibitem{NIPS2016_6393}
Bilen, H., Vedaldi, A.: Integrated perception with recurrent multi-task neural
  networks. In: Lee, D.D., Sugiyama, M., Luxburg, U.V., Guyon, I., Garnett, R.
  (eds.) Advances in Neural Information Processing Systems 29, pp. 235--243
  (2016)

\bibitem{clark2017}
Clark, T., Wong, A., Haider, M.A., Khalvati, F.: Fully deep convolutional
  neural networks for segmentation of the prostate gland in diffusion-weighted
  mr images. In: International Conference Image Analysis and Recognition. pp.
  97--104. Springer (2017)

\bibitem{Deng_ECCV_2014}
Deng, J., Ding, N., Jia, Y., Frome, A., Murphy, K., Bengio, S., Li, Y., Neven,
  H., Adam, H.: Large-scale object classification using label relation graphs.
  In: Fleet, D., Pajdla, T., Schiele, B., Tuytelaars, T. (eds.) Computer Vision
  -- ECCV 2014. pp. 48--64. Springer International Publishing, Cham (2014)

\bibitem{skin_cancer}
Esteva, A., Kuprel, B., Novoa, R.A., Ko, J., Swetter, S.M., Blau, H.M., Thrun,
  S.: Dermatologist-level classification of skin cancer with deep neural
  networks. Nature  \textbf{542},  115-- (Jan 2017),
  \url{http://dx.doi.org/10.1038/nature21056}

\bibitem{Fergus_ECCV_2010}
Fergus, R., Bernal, H., Weiss, Y., Torralba, A.: Semantic label sharing for
  learning with many categories. In: Daniilidis, K., Maragos, P., Paragios, N.
  (eds.) Proceedings of the European Conference on Computer Vision (ECCV). pp.
  762--775. Springer Berlin Heidelberg, Berlin, Heidelberg (2010)

\bibitem{goodfellow2014adversarialattack}
Goodfellow, I.J., Shlens, J., Szegedy, C.: Explaining and harnessing
  adversarial examples. arXiv preprint arXiv:1412.6572  (2014)

\bibitem{Guo_ICML_2017}
Guo, C., Pleiss, G., Sun, Y., Weinberger, K.Q.: On calibration of modern neural
  networks. In: Proceedings of the 34th International Conference on Machine
  Learning, {ICML} 2017, Sydney, NSW, Australia, 6-11 August 2017 (2017)

\bibitem{Hein_corr_2018}
Hein, M., Andriushchenko, M., Bitterwolf, J.: Why relu networks yield
  high-confidence predictions far away from the training data and how to
  mitigate the problem. CoRR  (2018)

\bibitem{huang2017densely}
Huang, G., Liu, Z., Van Der~Maaten, L., Weinberger, K.Q.: Densely connected
  convolutional networks. In: Proceedings of the IEEE conference on computer
  vision and pattern recognition. pp. 4700--4708 (2017)

\bibitem{huang2015bidirectional}
Huang, Z., Xu, W., Yu, K.: Bidirectional {LSTM-CRF} models for sequence
  tagging. arXiv preprint arXiv:1508.01991  (2015)

\bibitem{Kim_2018_CVPR}
Kim, E., Ahn, C., Oh, S.: {NestedNet:} learning nested sparse structures in
  deep neural networks. In: The IEEE Conference on Computer Vision and Pattern
  Recognition (CVPR) (June 2018)

\bibitem{Kokkinos_2017_CVPR}
Kokkinos, I.: Ubernet: Training a universal convolutional neural network for
  low-, mid-, and high-level vision using diverse datasets and limited memory.
  In: The IEEE Conference on Computer Vision and Pattern Recognition (CVPR)
  (July 2017)

\bibitem{cifar10}
Krizhevsky, A., Nair, V., Hinton, G.: Cifar-10 (canadian institute for advanced
  research) \url{http://www.cs.toronto.edu/~kriz/cifar.html}

\bibitem{cifar100}
Krizhevsky, A., Nair, V., Hinton, G.: Cifar-100 (canadian institute for
  advanced research) \url{http://www.cs.toronto.edu/~kriz/cifar.html}

\bibitem{kuncheva_2004}
Kuncheva, L.I.: Combining pattern classifiers: methods and algorithms. Wiley
  (2004)

\bibitem{lecun-mnisthandwrittendigit-2010}
LeCun, Y., Cortes, C.: {MNIST} handwritten digit database  (2010)

\bibitem{Liu_cvpr_2011}
Liu, B., Sadeghi, F., Tappen, M., Shamir, O., Liu, C.: Probabilistic label
  trees for efficient large scale image classification. pp. 843--850 (06 2013).
  \doi{10.1109/CVPR.2013.114}

\bibitem{luo2018towards}
Luo, P., Wang, X., Shao, W., Peng, Z.: Towards understanding regularization in
  batch normalization  (2018)

\bibitem{madry2017towards}
Madry, A., Makelov, A., Schmidt, L., Tsipras, D., Vladu, A.: Towards deep
  learning models resistant to adversarial attacks. arXiv preprint
  arXiv:1706.06083  (2017)

\bibitem{ipol_turbulence}
Meinhardt-Llopis, E., Micheli, M.: Implementation of the centroid method for
  the correction of turbulence. {Image Processing On Line}  \textbf{4},
  187--195 (2014). \doi{10.5201/ipol.2014.105}

\bibitem{plantvillage2016}
Mohanty, S.P., Hughes, D.P., Salath{\'e}, M.: Using deep learning for
  image-based plant disease detection. Frontiers in plant science  \textbf{7},
  ~1419 (2016)

\bibitem{moosavi2016deepfool}
Moosavi-Dezfooli, S.M., Fawzi, A., Frossard, P.: Deepfool: a simple and
  accurate method to fool deep neural networks. In: Proceedings of the IEEE
  conference on computer vision and pattern recognition. pp. 2574--2582 (2016)

\bibitem{Tishby_corr2_2017}
Moshkovitz, M., Tishby, N.: Mixing complexity and its applications to neural
  networks. CoRR  (2017)

\bibitem{naeini2015_AAAI}
Naeini, M.P., Cooper, G., Hauskrecht, M.: Obtaining well calibrated
  probabilities using bayesian binning. In: Twenty-Ninth AAAI Conference on
  Artificial Intelligence (2015)

\bibitem{Nguyen_CORR_2014}
Nguyen, A.M., Yosinski, J., Clune, J.: Deep neural networks are easily fooled:
  High confidence predictions for unrecognizable images. CoRR  (2014)

\bibitem{paninski2003estimation}
Paninski, L.: Estimation of entropy and mutual information. Neural computation
  \textbf{15}(6),  1191--1253 (2003)

\bibitem{papernot2017practical}
Papernot, N., McDaniel, P., Goodfellow, I., Jha, S., Celik, Z.B., Swami, A.:
  Practical black-box attacks against machine learning. In: Proceedings of the
  2017 ACM on Asia conference on computer and communications security. pp.
  506--519 (2017)

\bibitem{Parkhi_2015}
Parkhi, O.M., Vedaldi, A., Zisserman, A.: Deep face recognition. In: British
  Machine Vision Conference (2015)

\bibitem{pennington2014glove}
Pennington, J., Socher, R., Manning, C.D.: Glove: Global vectors for word
  representation. In: Proceedings of the 2014 conference on empirical methods
  in natural language processing (EMNLP). pp. 1532--1543 (2014)

\bibitem{ranjan2016hyperface}
Ranjan, R., Patel, V.M., Chellappa, R.: Hyperface: A deep multi-task learning
  framework for face detection, landmark localization, pose estimation, and
  gender recognition (2016)

\bibitem{Unet}
Ronneberger, O., P.Fischer, Brox, T.: U-net: Convolutional networks for
  biomedical image segmentation. In: Medical Image Computing and
  Computer-Assisted Intervention (MICCAI). LNCS, vol.~9351, pp. 234--241 (2015)

\bibitem{samangouei2018defense}
Samangouei, P., Kabkab, M., Chellappa, R.: Defense-gan: Protecting classifiers
  against adversarial attacks using generative models. arXiv preprint
  arXiv:1805.06605  (2018)

\bibitem{santurkar2018does}
Santurkar, S., Tsipras, D., Ilyas, A., Madry, A.: How does batch normalization
  help optimization? In: Advances in Neural Information Processing Systems. pp.
  2483--2493 (2018)

\bibitem{HL_fashion_classification}
Seo, Y., shik Shin, K.: Hierarchical convolutional neural networks for fashion
  image classification. Expert Systems with Applications  \textbf{116},  328 --
  339 (2019). \doi{https://do.org/10.1016/j.eswa.2018.09.022}

\bibitem{Tishby_corr_2017}
Shwartz{-}Ziv, R., Tishby, N.: Opening the black box of deep neural networks
  via information. CoRR  (2017)

\bibitem{sinha2017certifying}
Sinha, A., Namkoong, H., Duchi, J.: Certifying some distributional robustness
  with principled adversarial training. arXiv preprint arXiv:1710.10571  (2017)

\bibitem{breast}
Spanhol, F., Soares~de Oliveira, L., Petitjean, C., Heutte, L.: Breast cancer
  histopathological image classification using convolutional neural networks
  (07 2016). \doi{10.1109/IJCNN.2016.7727519}

\bibitem{Srivastava_nips_2013}
Srivastava, N., Salakhutdinov, R.R.: Discriminative transfer learning with
  tree-based priors. In: Burges, C.J.C., Bottou, L., Welling, M., Ghahramani,
  Z., Weinberger, K.Q. (eds.) Advances in Neural Information Processing Systems
  26, pp. 2094--2102. Curran Associates, Inc. (2013)

\bibitem{szegedy2017inception}
Szegedy, C., Ioffe, S., Vanhoucke, V., Alemi, A.A.: Inception-v4,
  inception-resnet and the impact of residual connections on learning. In:
  Thirty-First AAAI Conference on Artificial Intelligence (2017)

\bibitem{Szegedy_ICLR_2014}
Szegedy, C., Zaremba, W., Sutskever, I., Bruna, J., Erhan, D., Goodfellow,
  I.J., Fergus, R.: Intriguing properties of neural networks. In: Proceedings
  of the 2nd International Conference on Learning Representations (ICLR),
  Banff, AB, Canada, April 14-16, 2014, Conference Track Proceedings (2014)

\bibitem{Tishby_1999}
Tishby, N., Pereira, F.C., Bialek, W.: The information bottleneck method. pp.
  368--377 (1999)

\bibitem{triguero_2016_JPR}
Triguero, I., Vens, C.: Labelling strategies for hierarchical multi-label
  classification techniques. Pattern Recognition  \textbf{56},  170--183 (2016)

\bibitem{pmlr-v80-wehrmann18a}
Wehrmann, J., Cerri, R., Barros, R.: Hierarchical multi-label classification
  networks. In: Dy, J., Krause, A. (eds.) Proceedings of the 35th International
  Conference on Machine Learning. Proceedings of Machine Learning Research,
  vol.~80, pp. 5075--5084 (2018)

\bibitem{xiao2017/online}
Xiao, H., Rasul, K., Vollgraf, R.: Fashion-mnist: a novel image dataset for
  benchmarking machine learning algorithms (2017)

\bibitem{HL_traffic_recognition}
{Xuehong Mao}, {Hijazi}, S., {Casas}, R., {Kaul}, P., {Kumar}, R., {Rowen}, C.:
  Hierarchical cnn for traffic sign recognition. In: 2016 IEEE Intelligent
  Vehicles Symposium (IV). pp. 130--135 (June 2016)

\bibitem{HDCNN}
Yan, Z., Jagadeesh, V., DeCoste, D., Di, W., Piramuthu, R.: {HD-CNN:}
  hierarchical deep convolutional neural network for image classification.
  Proceedings of the IEEE International Conference on Computer Vision (ICCV)
  (2015)

\bibitem{zadrozny2002transforming}
Zadrozny, B., Elkan, C.: Transforming classifier scores into accurate
  multiclass probability estimates. In: Proceedings of the eighth ACM SIGKDD
  international conference on Knowledge discovery and data mining. pp.
  694--699. ACM (2002)

\bibitem{Zhao_nips_2011}
Zhao, B., Li, F., Xing, E.P.: Large-scale category structure aware image
  categorization. In: Shawe-Taylor, J., Zemel, R.S., Bartlett, P.L., Pereira,
  F., Weinberger, K.Q. (eds.) Advances in Neural Information Processing Systems
  24, pp. 1251--1259. Curran Associates, Inc. (2011)

\bibitem{zweig_ICCV_2007}
{Zweig}, A., {Weinshall}, D.: Exploiting object hierarchy: Combining models
  from different category levels. In: 2007 IEEE 11th International Conference
  on Computer Vision. pp.~1--8 (Oct 2007). \doi{10.1109/ICCV.2007.4409064}

\end{thebibliography}

\iftrue

\newpage{}
\begin{center}
  \textbf{\Large{Supplementary Material}}  
\end{center}

\appendix

\appendix
\counterwithin{table}{section}
\counterwithin{figure}{section}
\section{Additional examples}\label{app:sec:more_examples}
\counterwithin{table}{section}
\counterwithin{figure}{section}
\begin{figure}[htp]
    \centering
    \includegraphics[width=0.9\textwidth]{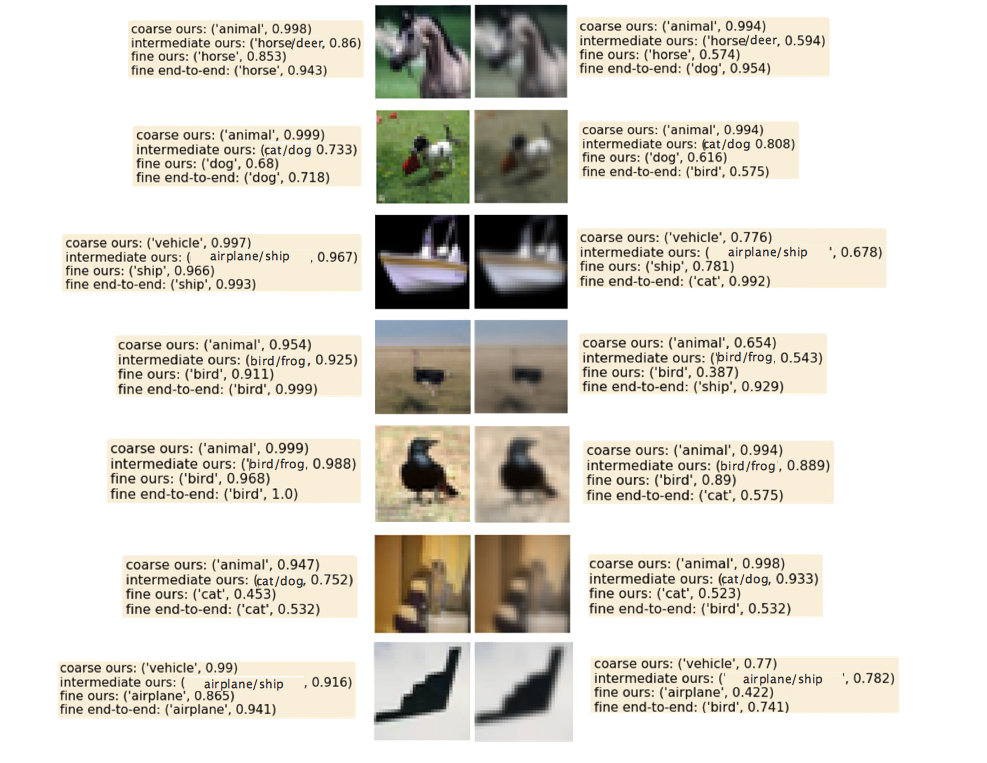}
    \caption{These results complement Figure~\ref{fig:example1}. The output of a standard (end-to-end) DNN and our proposed nested learning version are compared. 
    On the left, we show clean images from the test set of CIFAR-10 dataset, on the right, the same examples but blured. Next to each image our prediction (for the fine, middle, and coarse level) and the prediction of a standard (end-to-end) DNN are displayed. 
    Both DNNs share the same architecture and their performance is compared on Table~\ref{tab:cifar10_202020vs20} (rows  ``end-to-end, $|\mathcal{D}_3|=\frac{3N}{2}$'' and ``nested, $|\mathcal{D}_{1,2,3}|=N$''). 
    As shown in Table~\ref{tab:cifar10_202020vs20}, while the performance of both networks is similar on clean data (i.e., data that match the train distribution), learning nested representations significantly boost models robustness and mitigates overconfident predictions.}
    \label{app:fig:more_examples_usVSstandard}
\end{figure}

\section{Combination methods}\label{app:sec:combination}
\counterwithin{table}{section}
\counterwithin{figure}{section}
Combining multiple classifiers is a standard approach in machine learning. 
However, most approaches combine similar outputs and are not designed for the specific problem of nested learning. 
We compared standard combination methods and the proposed calibration-based combination strategy.

Combination methods can be classified into two categories. 
First those that combine classifiers predicted label, in this case, classifier outputs are considered as one-hot encoding vectors. 
For example, the \emph{Majority Vote (MA)} which consists in aggregating the decision of multiple classifiers and selecting the candidate that receives more votes.
The second category, combine classifiers continuous score outputs (rather than discrete labels).
As we show in the following, the second class of methods are more suitable for the combination of nested outputs. 

For example, given an input sample, a calibrated classifier provides an approximation of the probability associated to each coarse, middle, and fine labels.
In order to apply standard combination methods, we must map the coarse and intermediate predictions into a fine prediction and vice-versa. 
To this end, we assume coarse predictions have no information about the finer levels, and therefore, the probability of a coarse node is equally distributed across the fine nodes associated to it. 

Once probabilities associated to coarse levels are propagated to the fine levels and vice-versa, we can combine them using the \emph{mean} or the \emph{product} rule as described in \cite{kuncheva_2004}. 
Table~\ref{app:tab:combinationsMNIST101010} shows the result of combining nested outputs with standard combination methods and our strategy described in Section~\ref{subsec:CombinationOfOutputs}. 
\begin{table}[htp]
    \scriptsize
    \centering
    \begin{tabular}{|c|c|c|c|c|c|c|}
        \hline
        Distortion & Without comb.&  Ours & Coarse \& Fine& Mean &  Product & MV \\
        \hline
        Original     & \textbf{98.8} & 98.6 & 98.7 & 98.7 & 98.7 & 98.5 \\
        Distortion 2 & 80.8 & \textbf{82.8} & 82.1 & 82.2 & 82.4 & 80.1 \\
        Distortion 4 & 45.8 & \textbf{50.6} & 49.6 & 48.6 & 49.6 & 47.5  \\
      \hline
\end{tabular}
    \caption{\footnotesize Comparison of the fine accuracy for different combination techniques. The model is trained on MNIST dataset with $|\mathcal{D}_1|=|\mathcal{D}_2|=|\mathcal{D}_3|= 6000$ }
    \label{app:tab:combinationsMNIST101010}
\end{table}

\section{Implementation Details}\label{app:sec:implem_details}
\counterwithin{table}{section}
\counterwithin{figure}{section}
%
\subsection{Architecture}
\textbf{MNIST, Fashion-MNIST, CIFAR 10, Plantvillage.}
The architecture of our model is presented in Figure~\ref{app:fig:mnist_architecture}.
We used the same architecture for MNIST and Fashion-MNIST as the images have the same size and number of channels. 
For CIFAR10 and the Plantvillage data, the architecture of our model is very similar but with an increased depth of the convolutional filters.
We report the number of parameters for each model in Table~\ref{app:tab:params}.
Classifying images from CIFAR10 is indeed a harder problem than classifying images from MNIST or Fashion-MNIST, and therefore, it requires a model with more parameters.
\begin{table}[]
    \centering
    \scriptsize
    \begin{tabular}{|c||c|c|c|c|c|c|}
         \hline
         Dataset & MNIST & F-MNIST & Cifar 10& PlantVillage& DBPEDIA & CIFAR 100 \\
         \hline
         parameters& $5.2 \times10^4$ &$5.2 \times 10^4$&$ 6.3 \times 10^5$ &$2.4 \times 10^5$ & $4.2 \times 10^5 $ & $1.5 \times 10^6$\\
         \hline
    \end{tabular}
    \caption{Number of parameters for each nested model.}
    \label{app:tab:params}
\end{table}
The architecture we use is an adaptation of the U-Net \cite{Unet} designed to fit our proposed framework of nested information bottlenecks.  
The U-Net is indeed a good starting point, as it meets most of the criteria that we presented in Section \ref{sec:Theory}. 
First, it consists of a convolutional network that enforces a bottleneck representation. 
Second, it presents skip connections that allow information from the input to flow to deeper components of the network. 

We also added Batch-Normalization (BN) layers after every convolutional filters in order to achieve a faster convergence. BN is very helpful to mitigate the vanishing gradient phenomenon. Also, since it introduces randomness during training, BN acts as a regularization. Classic L1 regularization was shown to be similar to BN in \cite{luo2018towards}, while including BN layers also improves training speed and stability as explained by \cite{santurkar2018does}.   
\begin{figure}[t]
    \centering
    \includegraphics[width = 0.6\textwidth]{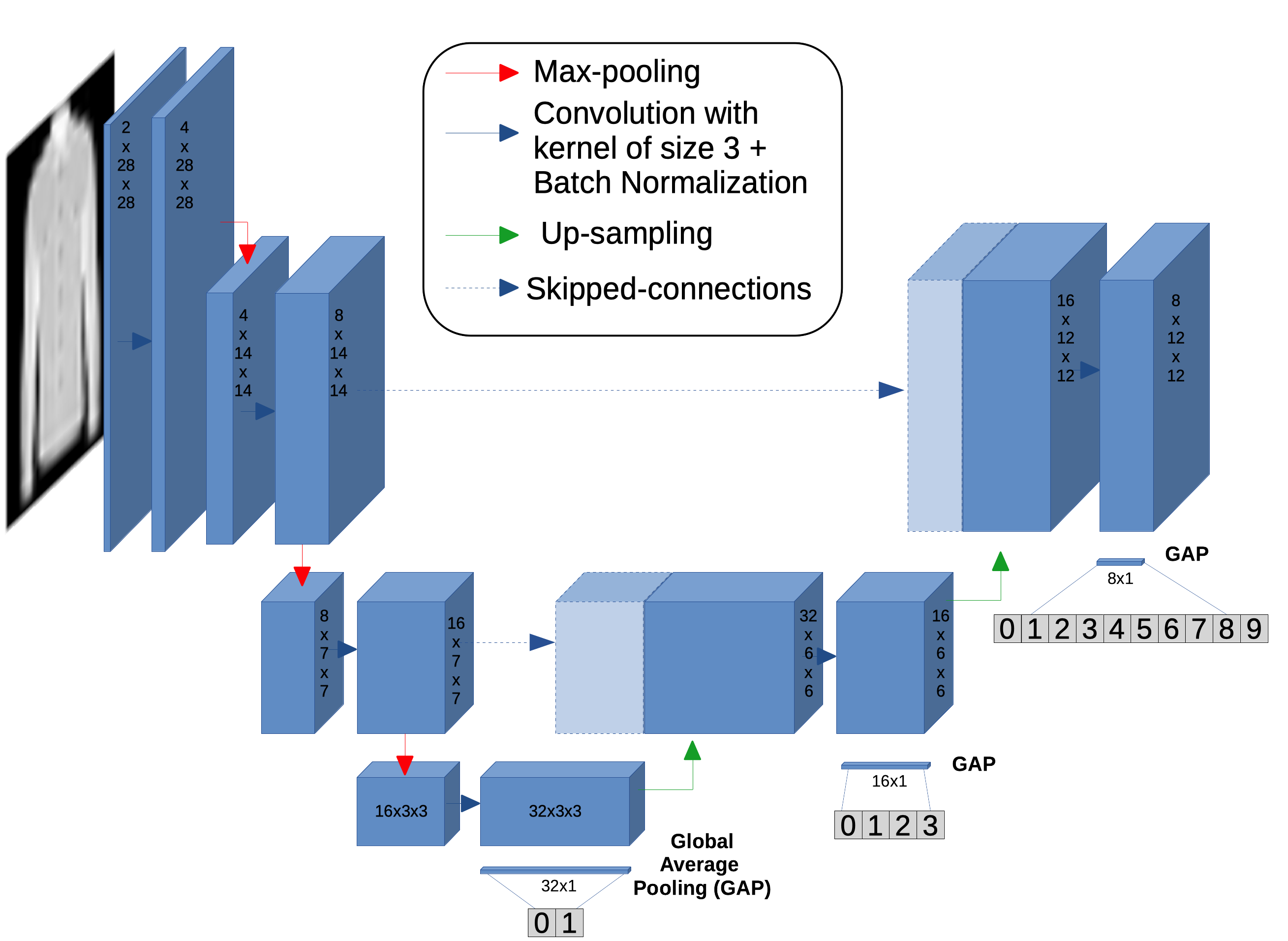}
    \caption{\footnotesize Architecture of our CNN for both Fashion-MNIST and MNIST. 
    This model is an adaptation of the U-Net network \cite{Unet} designed to fit our nested learning framework. 
    The blue boxes represent the features extracted by convolutional layers. 
    We perform a global average pooling rather than a flattening to decrease the number of parameters. 
    After the global average pooling, the feature vector is normalized with respect to the L2 norm (instance normalization layer).
    The normalized features are followed by fully connected layers to compute the final output. 
    The model used to test CIFAR10 set is very similar but has convolutional layers with more kernels to handle this (sightly more complex) task.}\label{app:fig:mnist_architecture}
\end{figure}

\noindent\textbf{DBPEDIA.} The architecture for the text data is presented in Figure \ref{app:fig:nlp_architecture_framework}. It is made of three important blocks: a preprocessing operation, a convolutional core, and task-specific layers. The preprocessing consists of a classic data cleaning and lemmatization. It is followed by an embedding to a (200,100) matrix using GloVe \cite{pennington2014glove}. Each line of this matrix corresponds to a vector word embedding of size 100. We pad or truncate articles to a fixed length of 200 to ease the implementation. This embedding is passed to a convotutional core, shown in yellow in Figure \ref{app:fig:nlp_architecture_framework}, which is similar to the U-Net, except that it is one dimensional. 
The intermediate features are passed to the task specific layers. Those layers are made of Bi-directional LSTMS and fully connected layers. Bi-directional LSTMs were shown to be particularly efficient for text classification and other Natural Language Processing tasks \cite{huang2015bidirectional}.
\begin{figure}[htp]
    \centering
    \includegraphics[width = 0.9\textwidth]{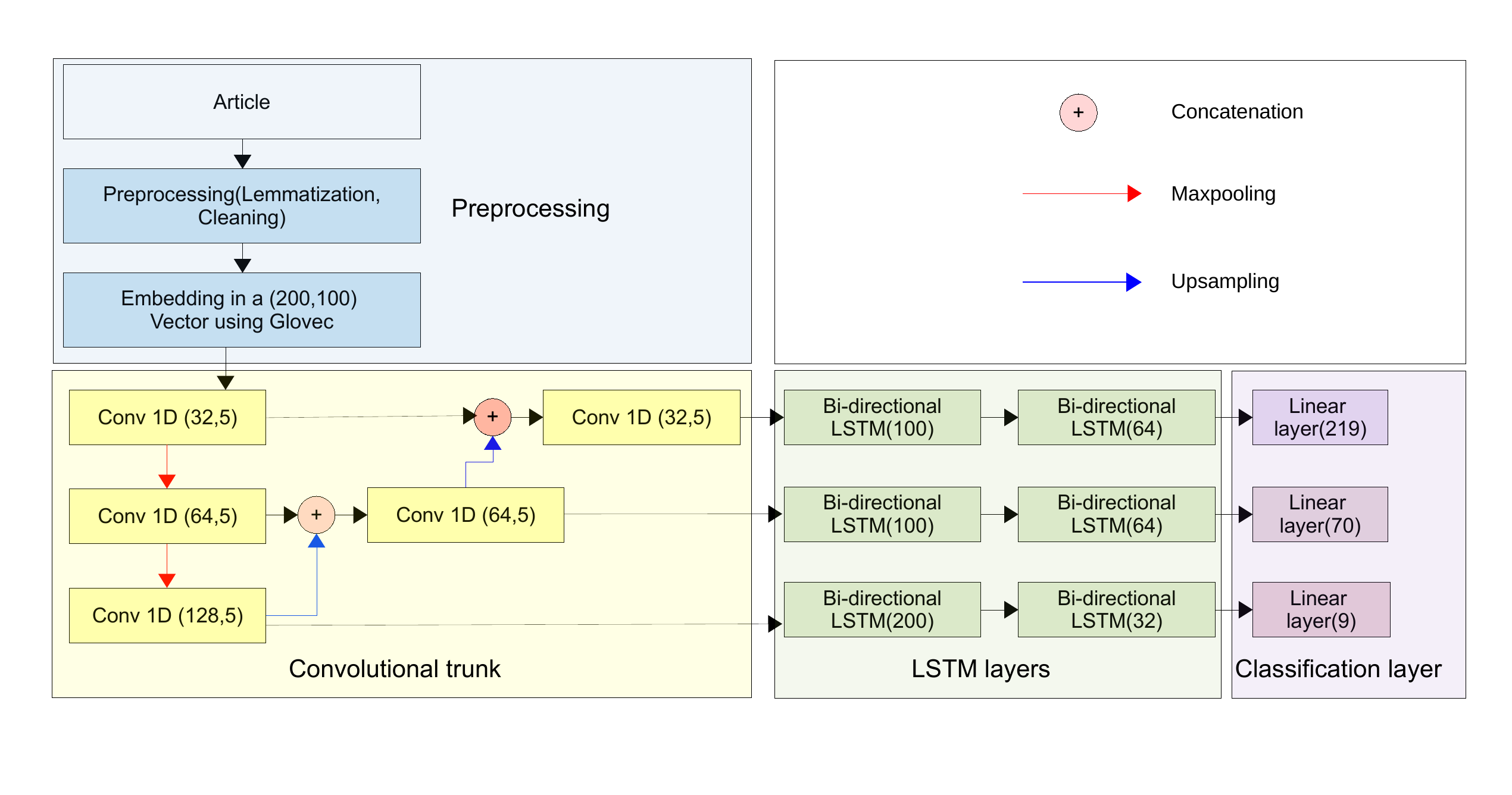}
    \caption{Preprocessing and architecture for the DBPEDIA model. The convolutional trunk is very similar to the one of a U-Net \cite{Unet}. It is associated to LSTM layers which are particularly suited for text data. The classification layers are fully connected layers. }
    \label{app:fig:nlp_architecture_framework}
\end{figure}

\noindent\textbf{CIFAR-100.} The architecture for the CIFAR100 dataset is presented in Figure~\ref{app:fig:cifar100_architecture}. This architecture is inspired by both the U-Net \cite{Unet} and the Densenet \cite{huang2017densely}. Dense net takes a step further in using residual connections by creating densely connected blocks. Those blocks are a sequence of convolution+batch normalization+ReLU layers applied to a concatenation of all the previously extracted features.
To get a coarse prediction we use the classic form of a densenet with three dense blocks, a global average pooling layer and a fully connected layer with 20 outputs . Then we upsample the features encoded by the last dense block and concatenate them with the features encoded by the previous dense block. These feature maps are then passed through a dense block which leads to a second average pooling layer and a fully connected layer of 100 outputs.
In our architecture, each dense block has the same parameters: 12 layers per block with a growth rate (depth) of 12, with bottleneck and compression layers (BC).  
\begin{figure}[htp]
    \centering
    \includegraphics[width = 0.6\textwidth]{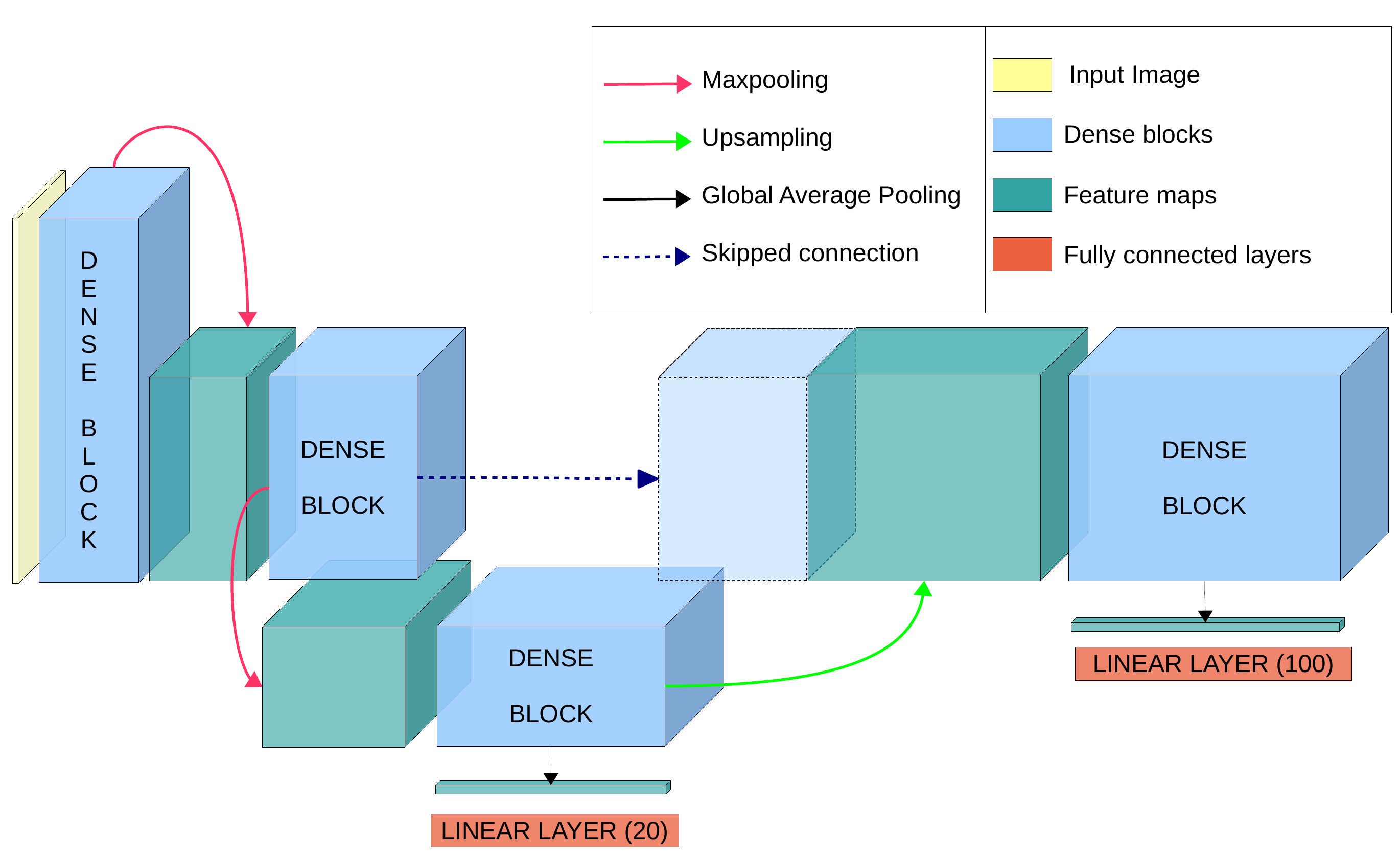}
    \caption{Architecture of our model for CIFAR-100 dataset.  This architecture is inspired by both the U-Net \cite{Unet} and the Densenet \cite{huang2017densely}.}
    \label{app:fig:cifar100_architecture}
\end{figure}

\subsection{Training}
We train the proposed nested models in an iterative fashion; First, we optimize the weights for the coarse prediction with the samples that are only coarsely annotated (freezing the remaining weights).
Then, we optimize the weights up to the intermediate output with the training samples with coarse and intermediate annotations.
This process is repeated for each level until we reach the finer level and the entire network is trained. 
Each training step is performed using ADAM optimizer with a specific learning rate. For example, for problems with three nested levels we set the learning rate as: $2\times10^{-3}$ for the first step, $1\times10^{-3}$ for the second, and $5\times10^{-4}$ for the final step.
We stop the training of each level when stagnation of the validation loss is observed.

\subsection{Calibration} 
The calibration consists of two main steps. First a ``rejection'' class is modeled using an uniform distribution on the latent space. 
This models out-of-distribution samples and mitigates overconfident predictions on portions of the feature space where no training data is observed. 
The second step consists of temperature scaling to convert output scores into approximations of class probabilities. 
\\

\noindent\textbf{The ``rejection'' class.} 
For every level of granularity $i$ and for every sample of the training dataset, we store the normalized outputs of the global averaging layer (GAP) in a dataset $D_i$. 
The samples have size $s_i$ and are normalized with respect to the L2 norm, therefore, they live in $\mathcal{B}^{s_i}(0,1)$, the unitary sphere in $\mathbb{R}^{s_i}$ centered at zero. 
We randomly sample $n_i$ new instances from an uniform distribution in $\mathcal{B}^{s_i}(0,1)$. These samples (associated to a new ``rejection'' class) are aggregated to $D_i$ and the fully connected layers fine tuned.\footnote{To this end, we used ADAM optimizer with a learning rate of $10^{-3}$.} 
Naturally, the larger $|D_i|$ and $s_i$, the larger $n_i$ should be.
We set $n_i \propto |D_i| \times \mathcal{S}(s_i)$, where $\mathcal{S}(s_i)$ is the area of the hypersphere of unitary radius in the $s_i$-dimensional space.  
Figure~\ref{app:fig:calib_toy} illustrates for a one dimensional toy example how the proposed ideas provide an efficient solution to reduce outputs overconfidence on out-of-distribution input samples.
\begin{figure}[h]
\centering     
\subfigure[]{\label{fig:a}\includegraphics[width = 0.32\textwidth]{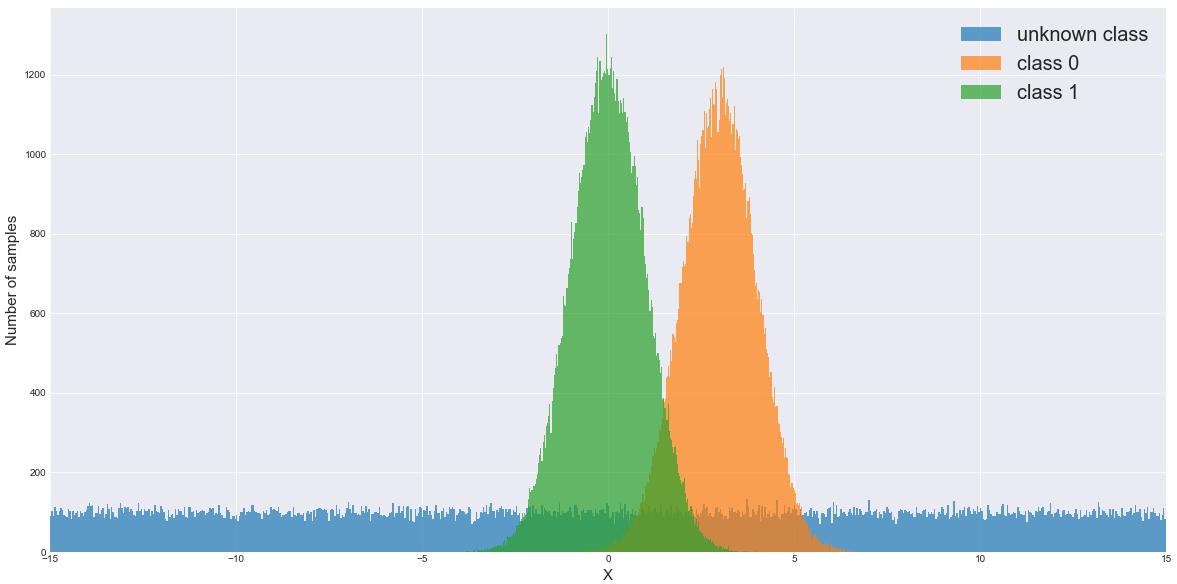}}
\subfigure[]{\label{fig:b}\includegraphics[width = 0.32\textwidth]{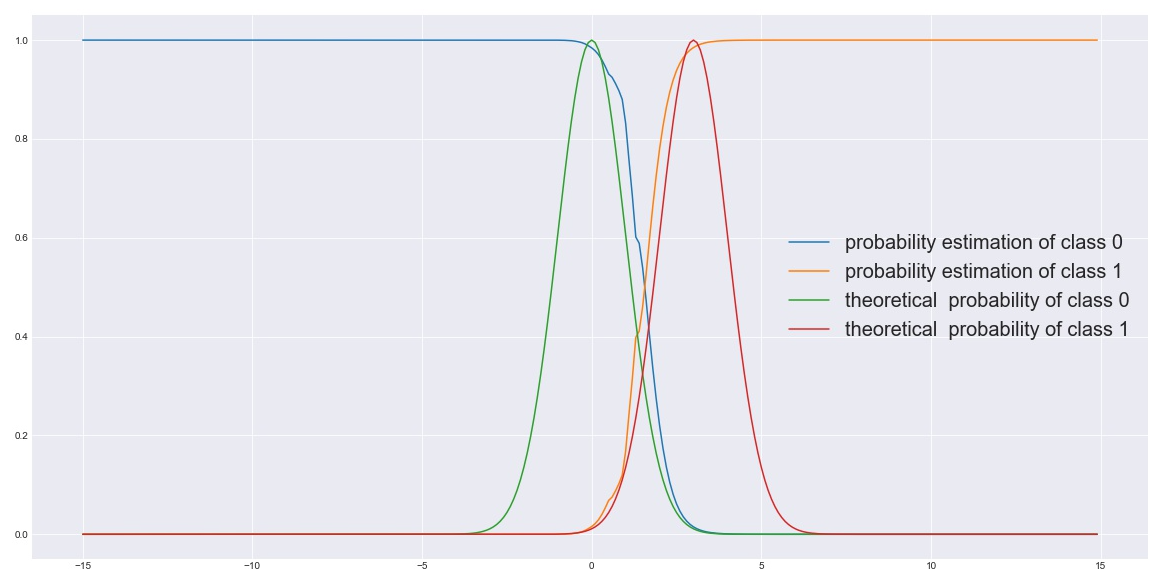}}
\subfigure[]{\label{fig:c}\includegraphics[width = 0.32\textwidth]{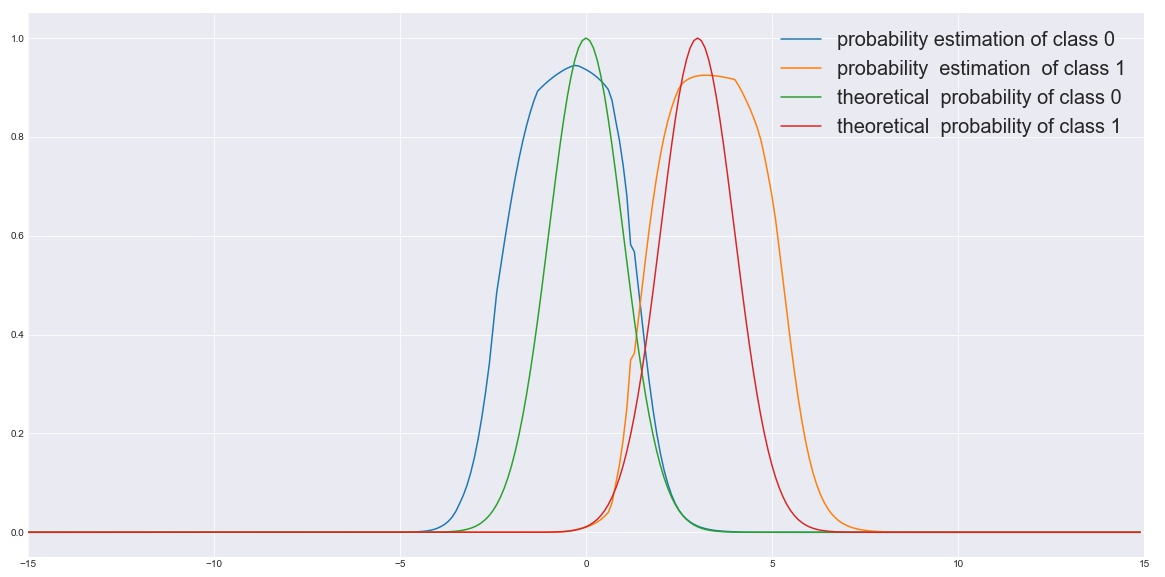}}
\caption{\footnotesize One dimensional toy example that illustrates the problem of prediction overconfidence for input samples that are far from the training distribution. 
In this example, two classes (``1'' and ``0'') are considered and we assume that the input $x\sim X$ is  one-dimensional. (a) illustrates the empirical distribution of each class (on green $P(X|Y=1)$ and yellow $P(X|Y=0)$).
In addition, we illustrate (blue distribution) the uniform distribution from which we sample synthetic training instances associated to the ``rejection'' class. 
Figure (b) shows the confidence output associated to the class ``1'' and ``0'' for different values of $x$, for a model trained only on the original data (standard approach). 
Figure (c) illustrates the output of the same DNN trained with the samples associated to the classes ``1'' and ``0'' plus the synthetic samples from the ``rejection'' class.}\label{app:fig:calib_toy}
\end{figure}
\\

\noindent\textbf{Temperature scaling.} 
Temperature scaling improves calibration by converting arbitrary score values into an approximation of class probabilities. 
As explained in Section \ref{subsec:Scores_calibration}, we  minimize the empirical ECE metric over the validation set, composed of 5000 unseen samples for all the dataset we tested our method on.
To find the optimal temperature $T$, we compute the ECE over $50$ values between $1$ and $3$ and select the one for which the ECE is minimized. 
If $T=1$ the model is already well calibrated. 
On all our experiments, the minimal ECE value was always reached for $T$ values strictly lower than 3. 
\section{Dataset taxonomies}\label{app:sec:label_grouping}
\counterwithin{table}{section}
\counterwithin{figure}{section}
\noindent{\textbf{Creating an artificial taxonomy for MNIST, Fashion-MNIST, CIFAR-10.}}
%
To group fine labels into a meaningful nested structure, we first train a shallow neural network and classify images into the fine classes. 
Then we used the confusion matrix $M$ associated to this auxiliary classifier to establish which classes are closer to each other. 
Note that this is done here for illustrating our proposed nested framework, other datasets already provide a natural taxonomy.
This way the levels of the nested structure can also have semantic meaning. But as demonstrated in the paper, even without explicit prior taxonomy, this type of nested structure improves performance for all levels.

For MNIST and Fashion-MNIST for example, we wanted to group the labels in $2$ coarse categories which also contained $2$ intermediate categories. 
To this end, we find the permutation $\sigma$ applied to both the rows and columns of $M$, such that the non-diagonal $5\times5$ matrices of $M$ had the lowest possible L1-norm. 
We iterate this process to find the intermediate categories.
It is computationally hard to go through all the permutations, therefore, we follow the ideas proposed in \cite{Behrisch:2016:MRM:3071534.3071595} to perform matrix reordering with a reduced complexity.

Figure \ref{app:fig:taxonomies} presents the groups of labels we obtained for MNIST, Fashion-MNIST, and CIFAR10. 
Our results show natural and intuitive taxonomies, see e.g., how MNIST digits are grouped according to a natural shape-oriented similarity, with 3 and 8 in the same intermediate class for example. A vertical left occlusion will force a standard DNN to pick between the two (or make a mistake outside this group), while obviously the correct answer is ``a 3 or an 8'' since there is no information for the fine class.
\begin{figure}[h]
\centering     
\subfigure[Taxonomy obtained for MNIST]{\includegraphics[width = 0.32\textwidth]{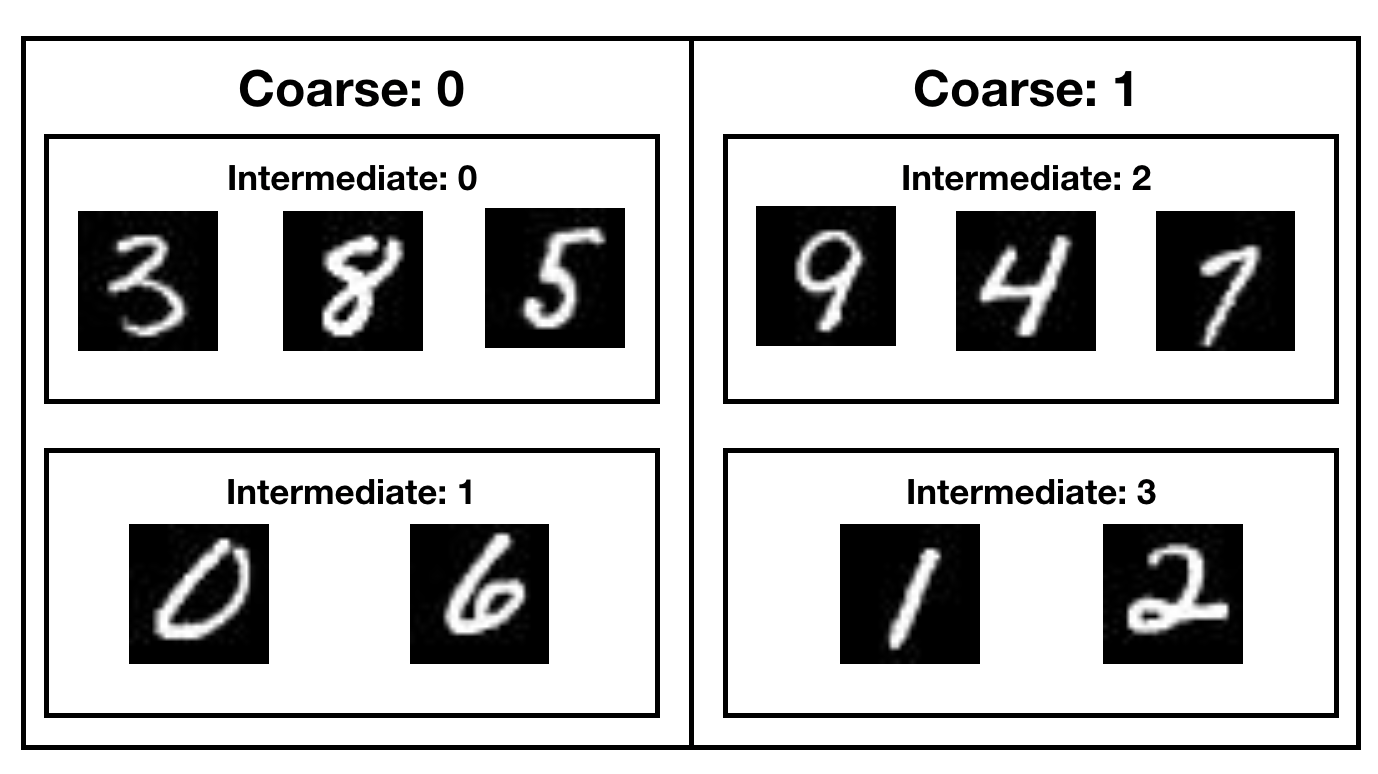}}
\subfigure[Taxonomy obtained for Fashion-MNIST]{\includegraphics[width = 0.32\textwidth]{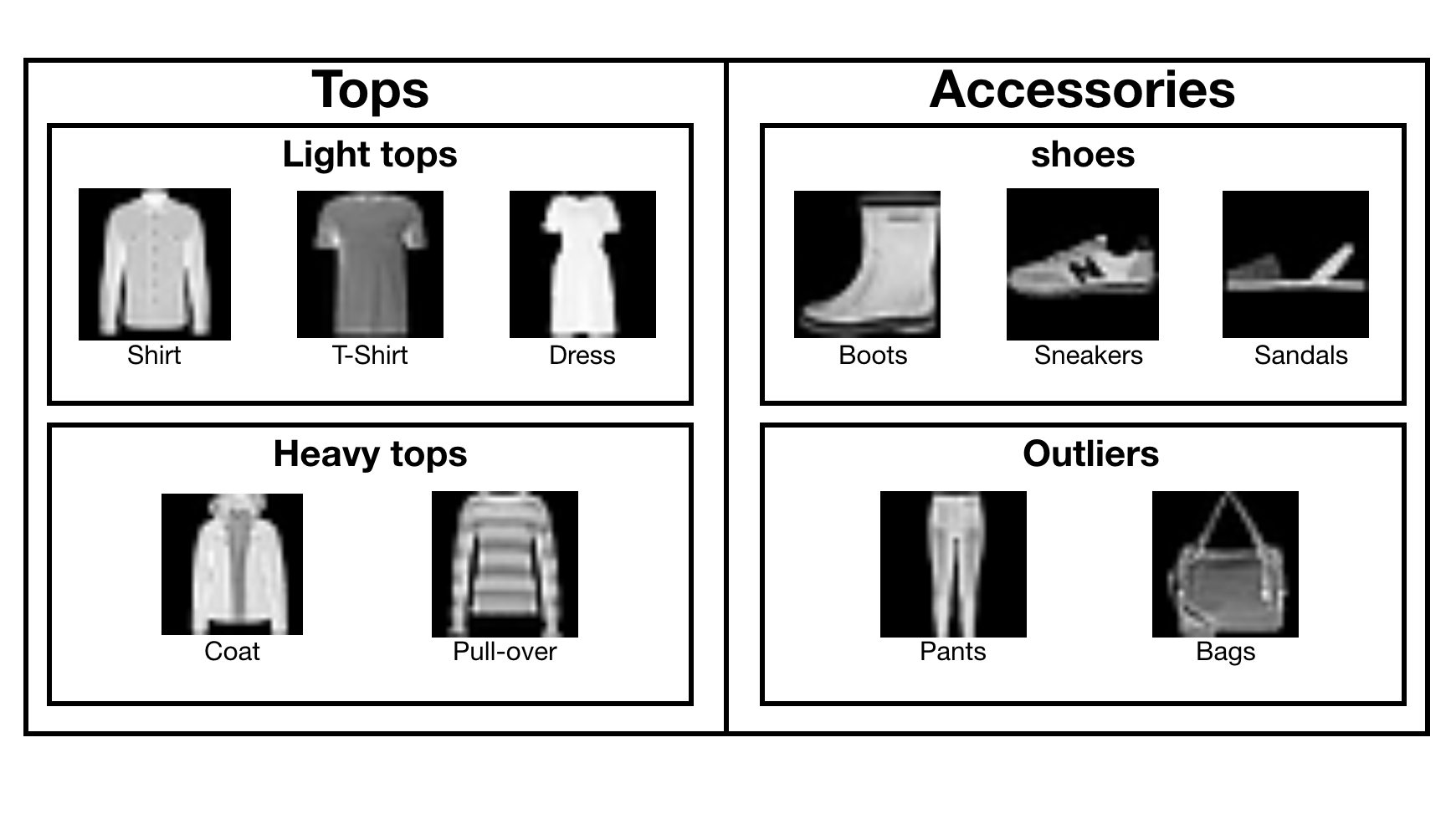}}
\subfigure[Taxonomy obtained for CIFAR10]{\includegraphics[width = 0.32\textwidth]{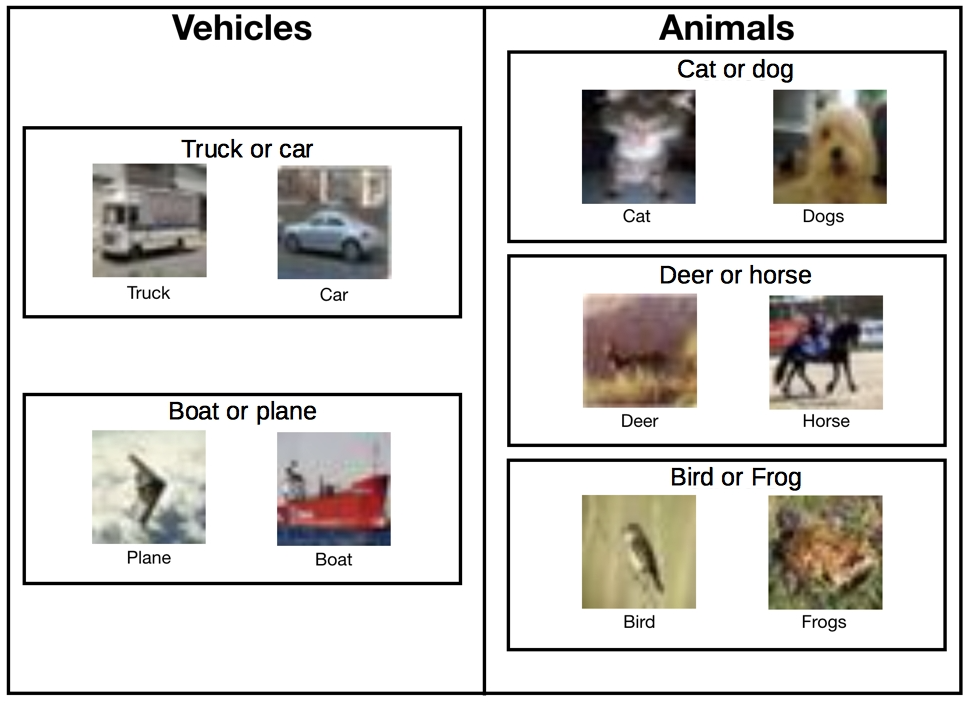}
}
\caption{\footnotesize Nested groups obtained by minimizing the non-diagonal components on the confusion matrix of an auxiliary simple classifier.}\label{app:fig:taxonomies}
\end{figure}

\noindent{\textbf{Taxonomies for CIFAR-100, DBPEDIA, and Plantvillage.}} 
Those datasets are already annotated following a nested taxonomy. CIFAR 100 is a dataset of small natural images of different objects. The labels are organized in 20 coarse classes (e.g., flower), each of them corresponding to 5 fine classes (e.g., orchids, poppies, roses, sunflowers, and tulips).
DBPEDIA is a large dataset of articles of wikipedia organized in a three level nested taxonomy. There are 9 coarse classes, 70 middle classes, 219 fine classes. For instance, an article of the Mugdock Castle near Glasgow, is labeled with the triplet (Place, Building, Castle).
Finally, Plantvillage dataset contains images of healthy and sick leaves from different species. The labels are not multi-granular. However the labels can be split in a natural nested taxonomy. In our framework, the coarse label corresponds to the species of the leaf ; the intermediate label corresponds to the binary label: healthy or sick . Finally, the fine labels corresponds to what sickness the plant has. Here are some examples of possible labels : (Apple, Apple healthy, Apple healthy), (Tomato, Tomato sick, Tomato Bacterial Spot), etc. By redefining the annotations in that fashion, we have 14 coarse classes, 23 middle classes, and 38 fine classes.

\newpage
\section{Additional experiments}\label{app:sec:additional_experiments}
\counterwithin{table}{section}
\counterwithin{figure}{section}
\subsection{Additional results for Fashion MNIST and MNIST}
\begin{table}[ht]
\begin{center}
\hspace{-5mm}
    \scriptsize
    \begin{tabular}{|c|c|c|c|c|c|c|}
    \hline
        ind & method & Original & Distortion 1 &  Distortion 2 &  Distortion 3 &  Distortion 4 \\
          \hline
          \hline
        \multicolumn{7}{|c|}{F-MNIST}\\
          \hline
        J& Coarse (end-to-end), $|\mathcal{D}_3| = \frac{3N}{2}$ & 98.9 / 99.3 & 90.7 / 96.4 &          87.5 / 95.9 &          85.1 / 95.5&          84.0 / 94.7 \\
         K & Coarse (end-to-end), $|\mathcal{D}_3| = N$ &          99.0 / 99.3 &94.5 / 96.9 &          90.1 / 96.1 &          87.3 / 94.9 &      \textbf{85.9} / 94.5 \\
          L  & Coarse (nested), $|\mathcal{D}_{1,2,3}| = N$ & \textbf{99.2} / 99.4 & \textbf{96.2} / 97.6 &          \textbf{93.1} / 96.6 & \textbf{89.3} / 96.0 &      84.7 / 95.5  \\
          \hline
         M& Middle (end-to-end), $|\mathcal{D}_3| = \frac{3N}{2}$  & \textbf{94.3} / 95.8  &82.1 / 90.9&          77.2 / 89.5&          73.0 / 88.2&          69.6 / 87.1 \\   
         N & Middle (end-to-end), $|\mathcal{D}_3| = N$  &          93.7 / 95.4 & 84.3 / 91.7 &          79.5 / 90.2 &          74.0 / 89.0 &      71.3 / 88.3 \\
          O  & Middle (nested), $|\mathcal{D}_{1,2,3}| = N$ &          94.2 / 94.7 &\textbf{87.3} / 91.2 &          \textbf{82.6} / 89.3 &          \textbf{77.7} / 88.5 &      \textbf{72.8} / 87.3  \\
          \hline
          P& Fine (end-to-end), $|\mathcal{D}_3| = \frac{3N}{2}$          & \textbf{88.3} / 91.5 &70.8 / 83.5 &          62.6 / 81.3 &          55.0 / 80.5 &          48.3 / 79.6     \\
          Q & Fine (end-to-end), $|\mathcal{D}_3| = N$ &          87.1 / 91.1 & 70.1 / 84.7 &          62.3 / 83.0 &          54.7 / 81.1 &      50.3 / 80.2 \\
            R &  Fine (nested), $|\mathcal{D}_{1,2,3}| = N$ &          86.7 / 87.9 &
            \textbf{73.7} / 81.3 &          \textbf{67.6} / 77.6 &          \textbf{57.6} / 76.4 &      \textbf{51.2} / 74.4  \\
          \hline
          \hline
        \multicolumn{7}{|c|}{MNIST}\\
          \hline
            S & Coarse (end-to-end), $|\mathcal{D}_3| = \frac{3N}{2}$  & 99.4 / 99.5 &   96.1 / 97.8 &          90.9 / 95.3 &          82.3 / 92.6 &          72.0 / 90.9   \\
            T & Coarse (nested), $|\mathcal{D}_{1,2,3}| = N$ &   \textbf{99.5} / 99.5 & \textbf{96.9} /  97.5&          \textbf{92.3} / 95.4&          \textbf{85.9} / 93.8&          \textbf{79.6} / 92.1   \\
            \hline
            U & Middle (end-to-end), $|\mathcal{D}_3| = \frac{3N}{2}$ & 99.1 /  99.2 & 94.8 / 96.9&          87.5 / 93.3&          74.6 / 89.8 &          60.7 / 88.0    \\
            V & Middle(nested), $|\mathcal{D}_{1,2,3}| = N$  & \textbf{99.2} / 99.2   & \textbf{95.5} / 96.0 &          \textbf{87.9} / 92.6 &          \textbf{77.0} / 89.4 &          \textbf{66.5} / 87.9 \\
            \hline
            W & Fine(end-to-end), $|\mathcal{D}_3| = \frac{3N}{2}$ &   98.5 / 98.7 &91.9 / 95.4&          80.4 / 90.9&          67.2 /  85.7 &          49.7 / 83.1      \\
            X &  Fine (nested), $|\mathcal{D}_{1,2,3}| = N$ &     \textbf{98.6}  / 98.4 &\textbf{92.8} /  93.5&          \textbf{82.8} / 87.8 &          \textbf{67.5} / 83.0  &          \textbf{50.6} / 80.4 \\
            \hline

\end{tabular}
\end{center}
    \caption{\footnotesize Complementary results of accuracy and mean confidence ($Acc\% / Conf\%$) for Fashion-MNIST, and MNIST datasets. In this experiments, we set $N = 6000$ Other results for Cifar 10, DBPEDIA, CIFAR100 and Plantvillage are reported in Table~\ref{tab:cifar10_202020vs20} and Table~\ref{tab:real_world_data}. }
    \label{app:tab:mnist_fmnist_202020vs20}
\end{table}

\vspace{-10mm}
\subsection{Comparing with a common MTL architecture}\label{app:sec:MTL}
%
A common Multi Task Learning (MTL) architecture consists of shared convolutional blocks followed by task-specific classification (fully connected) layers. 
Figure \ref{app:fig:new_architecture} illustrates an MTL architecture designed to obtain a nested classification of MNIST digits. 
The network enforces an information bottleneck that encodes the input into a (64,1) dimensional feature vector.
Then, it is connected to three classification branches implemented by a sequence of fully-connected layers.
This MTL model and our nested model have approximately the same number of parameters and are trained in an identical way on the MNIST dataset. 
\begin{figure}[htp]
    \centering
    \includegraphics[width = 0.8 \textwidth]{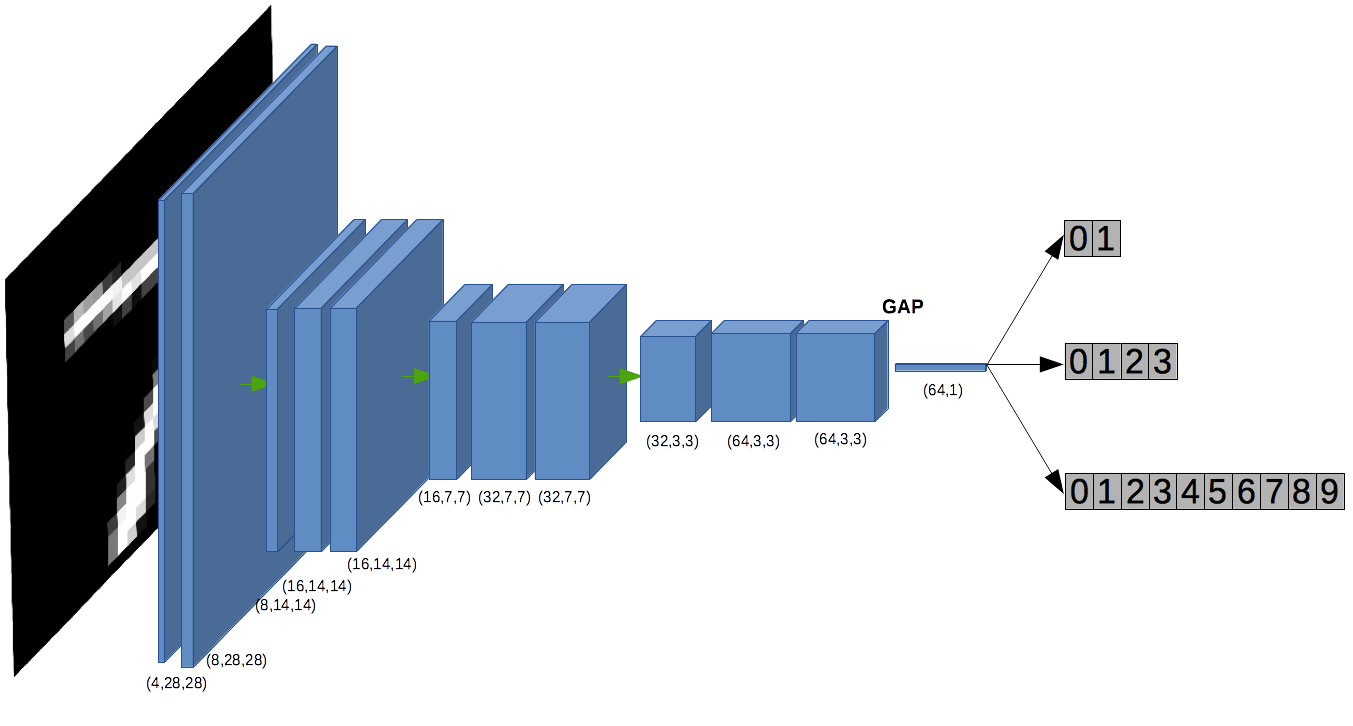}
    \caption{\footnotesize Architecture of the tested Multi-Task-Learning CNN. This architecture corresponds to a sequence of three blocks of two (convolutional + batch normalization) layers followed by a maxpooling layer. In order to reduce the dimension of the penultimate feature vector, we perform a global average pooling. This network and our nested CNN have approximately the same number of parameters and are trained using the same training protocol and data.}
    \label{app:fig:new_architecture}
\end{figure}
Table~\ref{app:tab:nested_vs_MTL} shows the classification accuracy for our nested architecture presented in Section \ref{app:sec:implem_details} and the Multi-Task Learning architecture presented above. 
On the original distribution both architectures perform roughly equivalently, but on perturbed test data the proposed nested model is considerably superior. 
Nested learning outperforms its MTL counterpart with a gap of 18\% for the fine task and a gap of more than 10\% for the coarse task for distortion levels 3 and 4. 
These results provide additional evidence that having a nested architecture promotes robustness and leads to more consistent models. 
\begin{table}[ht]
\scriptsize
\centering
        \begin{tabular}{|c|c|c|c|c|c|c|}
        \hline
                  Method & Original & Distortion 1 &  Distortion 2 &  Distortion 3 &  Distortion 4 \\
        \hline
        \hline
             Coarse (MTL),$|\mathcal{D}_{1,2,3}| = N$  & 99.5 / 99.7 &         92.4 / 96.5 &          82.6 / 94.3 &          73.8 /92.9 &          66.1 / 93.1      \\
             Coarse (nested) Ours, $|\mathcal{D}_{1,2,3}| = N$ &   \textbf{99.5} / 99.5 &       \textbf{96.9} /  97.5&          \textbf{92.3} / 95.4&          \textbf{85.9} / 93.8&          \textbf{79.6} / 92.1      \\
             \hline
             Middle (MTL), $|\mathcal{D}_{1,2,3}| = N$ & 99.1 /  99.4 &         91.1 / 94.9&          75.5 / 90.8&          60.2 / 88.8 &          49.9 / 88.2     \\
             Middle (nested) Ours, $|\mathcal{D}_{1,2,3}| = N$  & \textbf{99.2} / 99.2   &      \textbf{95.5} / 96.0 &          \textbf{87.9} / 92.6 &          \textbf{77.0} / 89.4 &          \textbf{66.5} / 87.9 \\
             \hline
             Fine(MTL), $|\mathcal{D}_{1,2,3}| = N$ &   \textbf{98.9} / 99.0 &       91.0 / 89.9&          70.9 / 81.3&          48.6 /  77.9 &          32.4/ 80.1      \\
              Fine (nested) Ours, $|\mathcal{D}_{1,2,3}| = N$ &     98.6  / 98.4 &     \textbf{92.8} /  93.5&          \textbf{82.8} / 87.8 &          \textbf{67.5} / 83.0  &          \textbf{50.6} / 80.4    \\
        \hline
        \end{tabular}
    \caption{\footnotesize Classification accuracy and mean confidence ($Acc\% / Conf\%$) for an example of MTL and nested learning on the MNIST dataset. Coarse, fine, and middle indicate the accuracy at each level of the label. MTL denotes the standard Multi-Task Learning architecture described in Section~\ref{app:sec:MTL}, while ``nested'' denotes the proposed nested model. In this experiment we set $N=12000$.}
    \label{app:tab:nested_vs_MTL}
\end{table}

\subsection{Skipped connections and their impact on the flow of information}\label{app:sec:skipedConnections}
Obtaining an empirical measure of the mutual information (MI) between two high dimensional random variables is a very hard numerical problem \cite{paninski2003estimation}.
However, recent progresses in deep learning made a numerical approximation (MINE algorithm) tractable by exploiting the flexibility of neural networks and properties of the mutual information. 
Belghazi et al.~\cite{belghazi2018mine} showed that a the problem of estimating MI, can be formulated as an optimization problem. 
They rely on the following characterization of the mutual information between random variables $X$ and $Z$  as the Kullback-Leibler (KL) divergence :
\begin{equation}
    I(X,Z) = D_{KL}(\mathbb{P}_{XZ} || \mathbb{P}_X \otimes \mathbb{P}_Z).
\end{equation}
Based on this, the authors use the Donsker-Varadhan representation of the KL divergence, which introduces a dual optimization problem,
\begin{equation}\label{app:eq:MI_dual}
    D_{KL}(\mathbb{P}||\mathbb{Q}) = \sup_{T:\Omega \rightarrow \mathbb{R}} \mathbb{E}_{\mathbb{P}}[T] - \log(\mathbb{E}_{\mathbb{Q}}[e^{T}]),
\end{equation}
where the supremum is taken over all the functions $T$ for which both terms in the right side of the equation are finite.

Since the parameters of a neural network can be used to encode a large space of functions, the authors propose to solve Equation~\ref{app:eq:MI_dual} for $T_\theta \in \mathcal{F}$, where $\mathcal{F}$ denotes the space of functions encoded by a pre-defined network architecture with parameters $\theta\in\Theta$. 
An approximation of the MI can be obtained solving the problem
\begin{equation}\label{app:eq:MINE}
    I_{\Theta}(X,Z) = \sup_{\theta \in \Theta} \mathbb{E}_{\mathbb{P}_{XZ}}[T_{\theta}] - \log(\mathbb{E}_{\mathbb{Q}}[e^{T_{\theta}}]).
\end{equation}
Equation~\ref{app:eq:MINE} can be solved in practice using standard optimization tools of deep learning (see \cite{belghazi2018mine} for details).

In order to measure the impact of skipped-connections from an MI perspective, we measure the empirical approximation of the MI between different sections of the proposed nested architecture. 
We compared these results for the same network with and without skipped connections.
Let us refer to the network with skipped connections with the subscript 1 and the network without skipped connections with the subscript 2.

We define different variables of interest at particular stages of the network and estimate the mutual information between them.
We will refer as $F_1(X)$ and $F_2(X)$ the variable corresponding to the feature map created after the 2nd maxpooling layer with and without skipped connections (see figures \ref{app:fig:mnist_architecture} and 4). 
Similarly, let $G_1(X)$ and $G_2(X)$ be the feature maps obtained before the second global average pooling (GAP) layer.
We will also consider $H_1(X)$ and $H_2(X)$ the coarsest feature maps obtained before the first GAP.
(Before feeding these features to the MINE algorithm we performed average pooling to reduce the dimension of the input variables.)

We estimated the MI between $I(F_i(X), G_i(X))$ and $I(F_i(X), H_i(X))$, $i=1$ meaning ``with skipped connections'' and $i=2$ ``without skipped connections''. (The network without skipped connections is re-trained to allow the model to adapt to this new configuration).  
%
%
Figure \ref{fig:MI} (left side) sketches in which sections of the model we are measuring the mutual information, and (right side) the evolution of the MI estimation (MINE algorithm) for $3000$ steps. 
\begin{align}\label{app:eq:results_mi}
    \Delta_1 = I(F_1(X),G_1(X))- I(F_1(X),H_1(X)) = 0.37, \\
    \Delta_2 = I(F_2(X),G_2(X))- I(F_2(X),H_2(X)) = 0.21 \approx{\Delta_1/2}.
\end{align}
As shown in equations 6 and 7 and discussed in Section 4, skipped connections play an important role to allow information to flow to the coarse feature representations of the finer representations. 
This provides additional numerical evidence to the discussion presented in Section~\ref{sec:Theory} supporting the importance of including skipped connections and avoiding information bottleneck at the finer classification. 

\section{Perturbations}\label{app:sec:perturbation}
\counterwithin{table}{section}
\counterwithin{figure}{section}
Selecting realistic and meaningful perturbations to test DNN models is a non trivial problem. 
For example, adding Gaussian noise mainly affects the high frequency components of the input images and we observed that both standard end-to-end and nested networks are not affected by this type of perturbation.  
In this work we focus on structural deformations inspired by a model of turbulence. 
The pseudo-code of this perturbation is presented in Algorithm \ref{app:algo1} and was inspired by \cite{ipol_turbulence}.
Figure~\ref{app:fig:examples_distortion} illustrates the distortion of an example image from the MNIST dataset for different levels of \emph{turbulence intensity}.
\begin{figure}[htp]
    \centering
    \includegraphics[width = 0.19\textwidth]{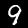}
    \includegraphics[width = 0.19\textwidth]{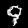}
    \includegraphics[width = 0.19\textwidth]{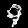}
    \includegraphics[width = 0.19\textwidth]{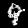}
    \includegraphics[width = 0.19\textwidth]{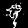}
    \caption{\footnotesize Example of different perturbations applied during testing time. From left to right: the original sample, and the distorted version with parameters $(S,T) = (1,0.8)$, $(S,T) = (1,1.0)$, $(S,T) = (1,1.3)$, and $(S,T) = (1,1.5)$ respectively.}
    \label{app:fig:examples_distortion}
\end{figure}
This perturbation is very helpful to test the robustness of neural networks as it affects the appearance, the edges, and morphology of the presented objects.  
\begin{algorithm}[htp]
\SetAlgoLined
\KwData{the input: $I\in \mathbb{R}^{w,h}$, the parameters $(S,T) \in \mathbb{R}^2$}
\KwResult{the distorted image $I_{dist}^{(S,T)}$}
Creating a vector field (u,v) for the distortion:\\
$u,v = normal\_noise((w,h)), normal\_noise((w,h))$\\
$u,v = gaussian\_filter(u,S), gaussian\_filter(v,S)$\\
$u,v = u\times \frac{T}{std(u)}, v\times \frac{T}{std(v)}$\\
Interpolate the image with the obtained vector field: \\
$I_{dist}^{(S,T)} = bilinear\_interpolate(I,u,v)$
\caption{Turbulence distortion}\label{app:algo1}
\end{algorithm}

\section{Geometry}\label{app:Geometry}
In addition to looking at the evolution of the accuracy during the cascaded training of our model in Section~\ref{sec:Experiments}, we also investigate the behaviour of some specific features of the network by visualizing their 2D-TSNE embedding.
This analysis is also extremely helpful to intuitively understand some of the results obtained in Section 5. 
For example, why our model is less sensitive to adversarial attacks, and why even when the fine label is misled the middle and coarse predictions remain rather accurate. 

We chose to visualize the embedding of the features extracted by the global average pooling layers of the nested MNIST model and the end-to-end model. 
We do so at each of the three main steps of our training method for the nested model (i.e. training with coarse annotations only first, training with coarse and middle annotations, and finally training on coarse middle and fine annotations) and after 50 and 100 epochs for the end-to-end model.

As we can see in Figure~\ref{app:fig:geometry}, the feature embedding of the nested model have a much more reasonable structure than the end-to-end model. 
This makes sense, as end-to-end learning tries to find a representation that separates different classes, but there is no explicit reward to map similar classes closer to each other.
On the other hand, nested learning encourages a nested representation as we can see in Figure G.1, from all the possible solutions that separate the fine classes, our model is choosing one that maps together fine classes with nested similarities.

%
If we look at the evolution of the feature embeddings, we can clearly identify clusters, which correspond to coarse middle and fine classes as we introduce more than one level of annotation, whereas the data points are almost distributed uniformly for the end-to-end model. 
Even though the accuracy of the fine predictions are similar, the feature embedding of our model shows a better and more natural 2D structure, with fine classes of the same middle class being close to each other. For example, the samples from classes 4, 7, and 9 -which belong to the same intermediate class- are close to each other in the 2-D embedding of the features extracted by the last GAP layer.
\begin{figure}
    \centering
    \includegraphics[width = \textwidth]{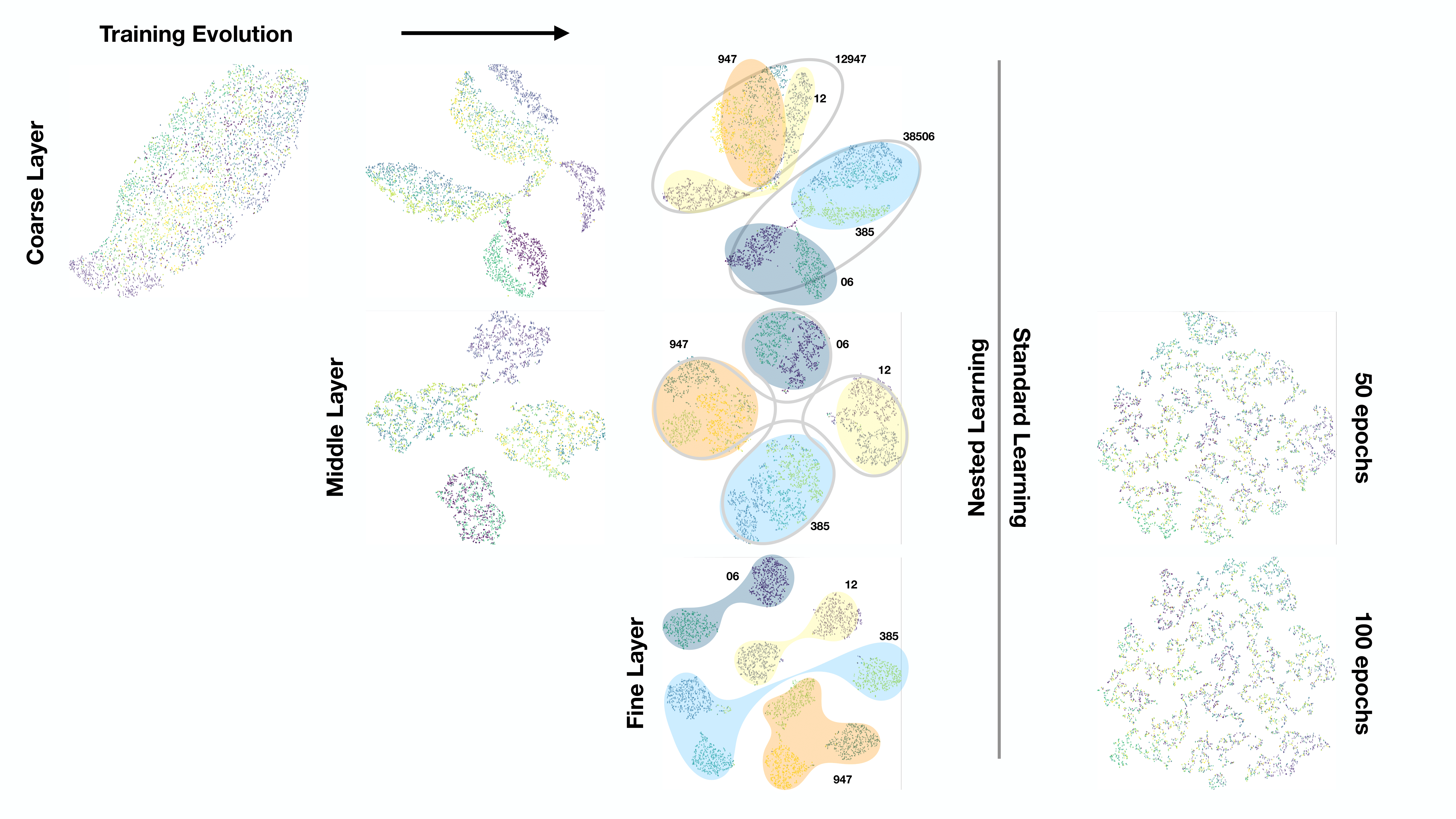}
    \caption{2D feature embedding for nested versus end-to-end learning. From left to right, we see how the feature embedding evolves as training proceeds. The first row shows the feature embedding of the features prior the coarse prediction, the middle row the embedding associated to the middle prediction, and finally, the last row the embedding associated to the fine prediction. On the right side, we compare the feature embedding for the fine prediction trained in an end-to-end fashion.}
    \label{app:fig:geometry}
\end{figure}

\fi
\end{document}